\RequirePackage{fix-cm}
\documentclass[smallextended, natbib]{svjour3}       
\smartqed  
\usepackage{float}
\usepackage[section]{placeins}
\usepackage{graphicx}
\usepackage[utf8]{inputenc}
\usepackage[T1]{fontenc}
\usepackage{subcaption}
\usepackage{url}
\usepackage[boxed,linesnumbered]{algorithm2e}
\usepackage{listings}
\usepackage{mathtools}
\usepackage{amsfonts}
\usepackage{amssymb}
\usepackage{hyperref}
\usepackage{color}
\def\CE{\textit{CE}}
\def\HF{\textit{HF}}
\def\HE{\textit{HE}}

\begin{document}

\title{Heuristics, Answer Set Programming and Markov Decision Process for Solving a Set of Spatial Puzzles\thanks{Thiago Freitas dos Santos is sponsored by FAPESP-IBM Proc. 17/07833-9. 
Paulo E. Santos acknowledges support from FAPESP-IBM Proc. 2016/18792-
9. 
Leonardo Anjoletto Ferreira acknowledges financial support from Coordena\c{c}\~{a}o de Aperfei\c{c}oamento de Pessoal de N\'{i}vel Superior -- Brasil (CAPES) -- Finance Code 001. Reinaldo Bianchi acknowledges financial support from FAPESP Proc. 2016/21047-3.}
}
\subtitle{}

\titlerunning{HoASP(MDP) for Solving a Set of Spatial Puzzles}        

\author{Thiago Freitas dos Santos        \and
        Paulo E. Santos \and
        Leonardo A. Ferreira \and
        Reinaldo A. C. Bianchi \and
        Pedro Cabalar.
}


\institute{Thiago Freitas dos Santos \at
              Centro Universitário FEI;
              \email{thiagosantos38@gmail.com}           
           \and
           Paulo E. Santos \at
              Centro Universitário FEI;
              \email{psantos@fei.edu.br}           
           \and
           Leonardo A. Ferreira \at
              Centro Universitário FEI;
              \email{laferreira@fei.edu.br}           
           \and
           Reinaldo A. C. Bianchi \at
              Centro Universitário FEI;
              \email{rbianchi@fei.edu.br}           
           \and
           Pedro Cabalar \at
              University of Corunna;
              \email{cabalar@udc.es} 
}

\date{Received: date / Accepted: date}
\maketitle
\begin{abstract}
Spatial puzzles composed of rigid objects, flexible strings and holes offer interesting domains for reasoning about spatial entities that are common in the human daily-life's activities. The goal of this work is to investigate the automated solution of this kind of puzzles adapting an algorithm that combines Answer Set Programming (ASP) with Markov Decision Process (MDP), algorithm oASP(MDP), to use heuristics accelerating the learning process. ASP is applied to represent the domain as an MDP, while a Reinforcement Learning algorithm (Q-Learning) is used to find the optimal policies. In this work, the heuristics were obtained from the solution of relaxed versions of the puzzles. Experiments were performed on deterministic, non-deterministic and non-stationary versions of the puzzles. Results show that the proposed approach can accelerate the learning process, presenting an advantage when compared to the non-heuristic versions of oASP(MDP) and Q-Learning. 
\end{abstract}

\section{Introduction}
\label{intro}

The capacity of learning, from domain interactions, actions and sequences of actions to solve complex tasks is an essential ability for any intelligent agent immersed in the physical human world. This is particularly critical with respect to spatial domains containing rigid, as well as flexible (or holed) objects, whereby the actions and their effects are non-trivial. This is the main challenge considered in this work, namely, learning sequences of actions necessary to solve a given task from the interaction with spatial domains. In this paper, the task of interest is finding solutions for a set of spatial puzzles composed of rigid objects, flexible strings and holes. Not only are these types of elements the composing parts of common human scenarios, but they are also of interest to application areas such as robot surgery and machine maintenance, in which objects with distinct (or contrasting) characteristics have to be carefully manipulated in order to achieve a particular goal (that could be the removal of a tumor, or the repair of a broken mechanism). 

Previous work has tackled the automated solution of such spatial domains from a logical or formal perspective~\citep{cabalar2011formalising,cabalar2016qualitative,santos2016framing}, where the actions and their effects were explicitly formalized, allowing the definition of a simple planning system capable of solving a number of such puzzles \citep{cabalar2011formalising}. These earlier approaches, however, did not have a learning component and could not cope with more complex and non-stationary versions of the puzzles.





In order to solve these issues, the present paper focuses on applying a combination of Answer Set Programming (ASP)~\citep{eiter2009answer} and Markov Decision Process (MDP)~\citep{Sutton-Barto2ed}, the oASP(MDP) algorithm (Section  \ref{oaspmdp}), to solve the Fisherman's Folly and Rope Ladder puzzles (Section \ref{background-Domain}). In our solution, ASP is used to represent the domain as an MDP, while Q-Learning is the Reinforcement Learning (RL) algorithm used to find an optimal policy for this MDP~\citep{ferreira2018method}. The main contribution of this work is the introduction of heuristics accelerating the learning procedure of the oASP(MDP) algorithm. These heuristics were obtained from solving simplified versions of the puzzles (Section \ref{method}) and they facilitated the solution of deterministic and non-deterministic versions of the domains, as well as versions where the string is allowed to wind around the holes, issues that were outside the scope of previous related work \citep{cabalar2011formalising,cabalar2016qualitative,santos2016framing}. 

Experiments (Section \ref{experiments}) and results (Section \ref{results}) show that our approach outperforms the non-heuristic version of the oASP(MDP) and Q-Learning algorithms.

\section{Background}
\label{background}
This section presents a description of the base concepts used in this work. Starting with the description of the spatial puzzles explored, we then present a brief description of Answer Set Programming (ASP), used to represent the Markov Decision Process (MDP), and the Reinforcement Learning technique used to find an optimal solution for this MDP. Finally, the non-heuristic version of the oASP(MDP) algorithm is presented (its heuristic version is described in Section \ref{method}).

\subsection{Domain}
\label{background-Domain}
This work considers two domains, the Fisherman's Folly and the Rope Ladder puzzles, that have the characteristics we want to investigate with respect to spatial problems, such as flexible and holed objects.

\begin{figure*}[t!]
\centering
\begin{subfigure}[t]{0.45\textwidth}
	\centering
	\includegraphics[width=\textwidth]{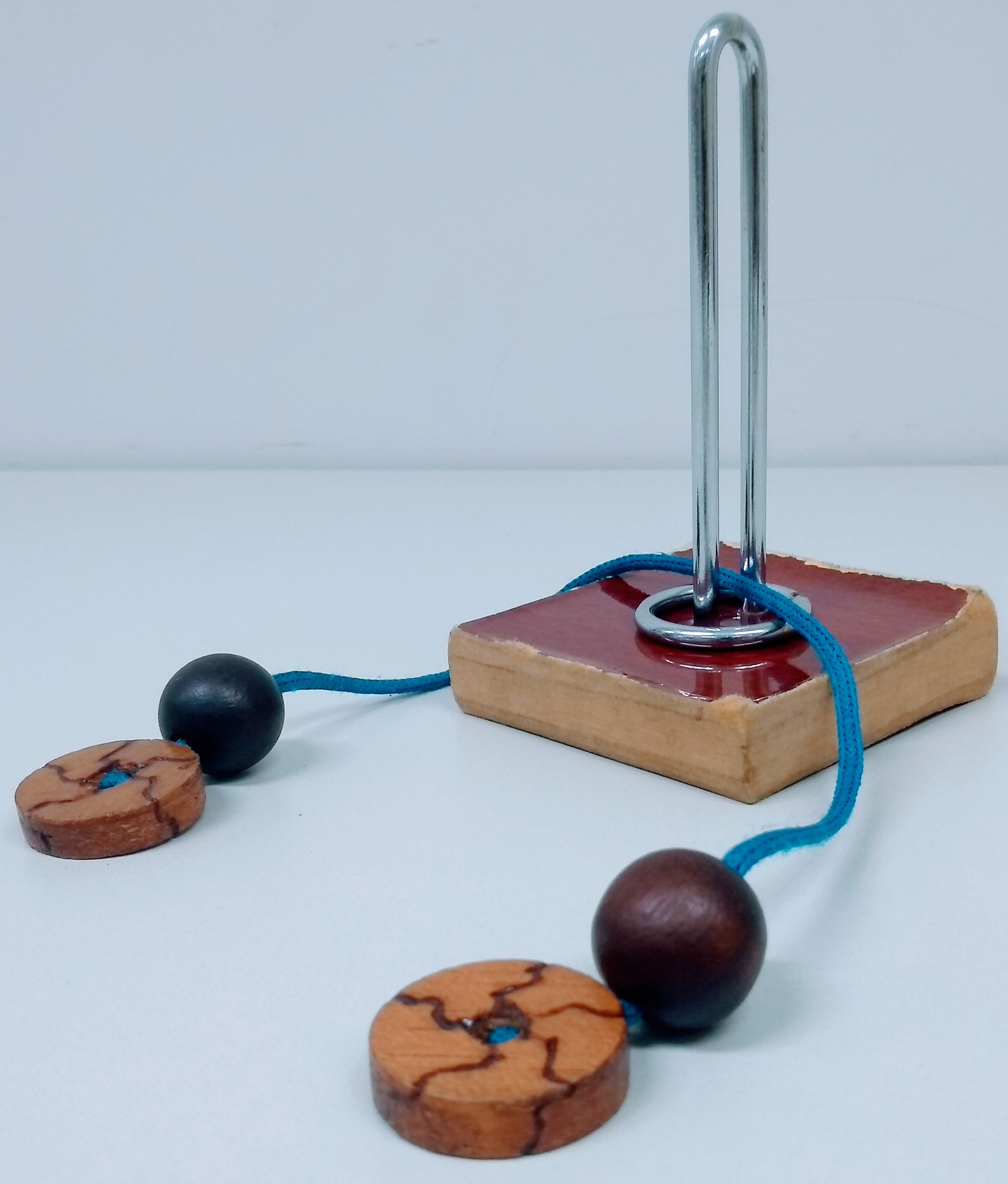}
	\caption{The Fisherman's Folly puzzle.}
	\label{fig:FishermansFollyPuzzle}
\end{subfigure}
\hfill
\begin{subfigure}[t]{0.45\textwidth}
	\centering
	\includegraphics[width=\textwidth]{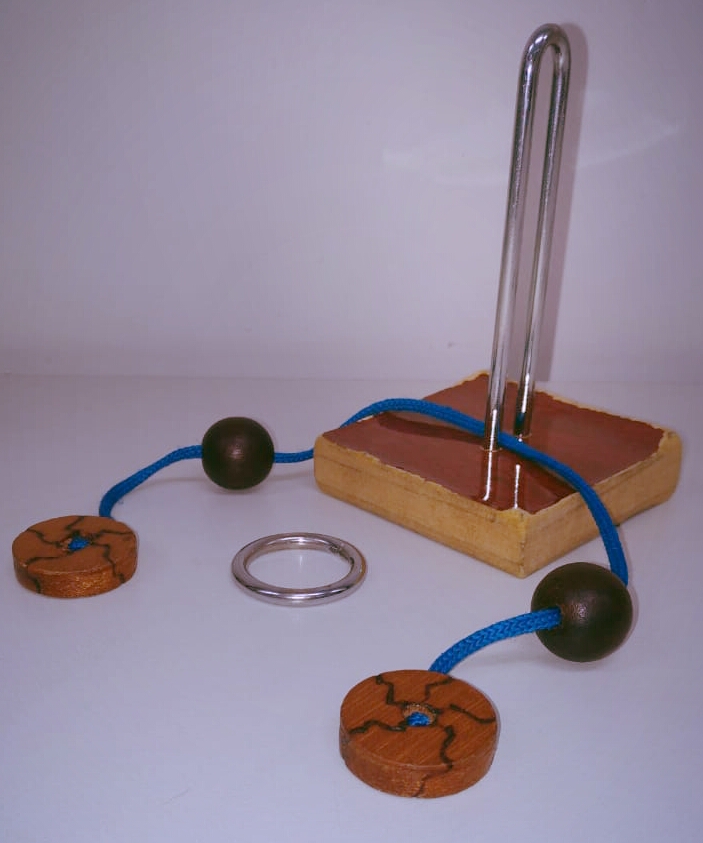}
	\caption{The goal of the Fisherman's Folly puzzle.}
	\label{fig:FishermansFollyPuzzleGoal}
\end{subfigure}
\\
\begin{subfigure}[t]{0.45\textwidth}
	\centering
	\includegraphics[width=\textwidth,height=1.05\textwidth]{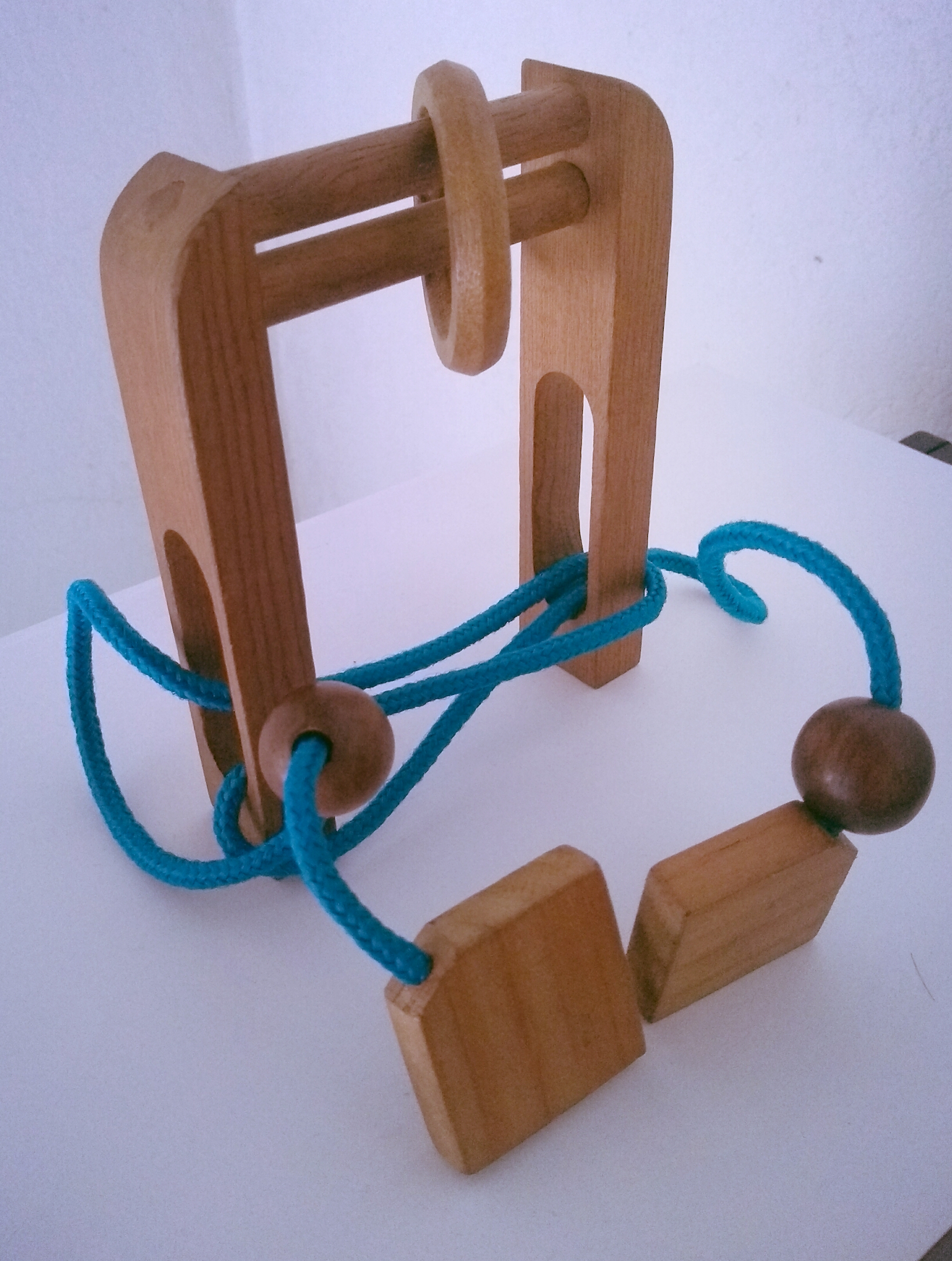}
	\caption{The Rope Ladder puzzle.}
	\label{fig:RopeLadderPuzzle}
\end{subfigure}
\hfill
\begin{subfigure}[t]{0.45\textwidth}
	\centering
	\includegraphics[width=\textwidth,height=1.05\textwidth]{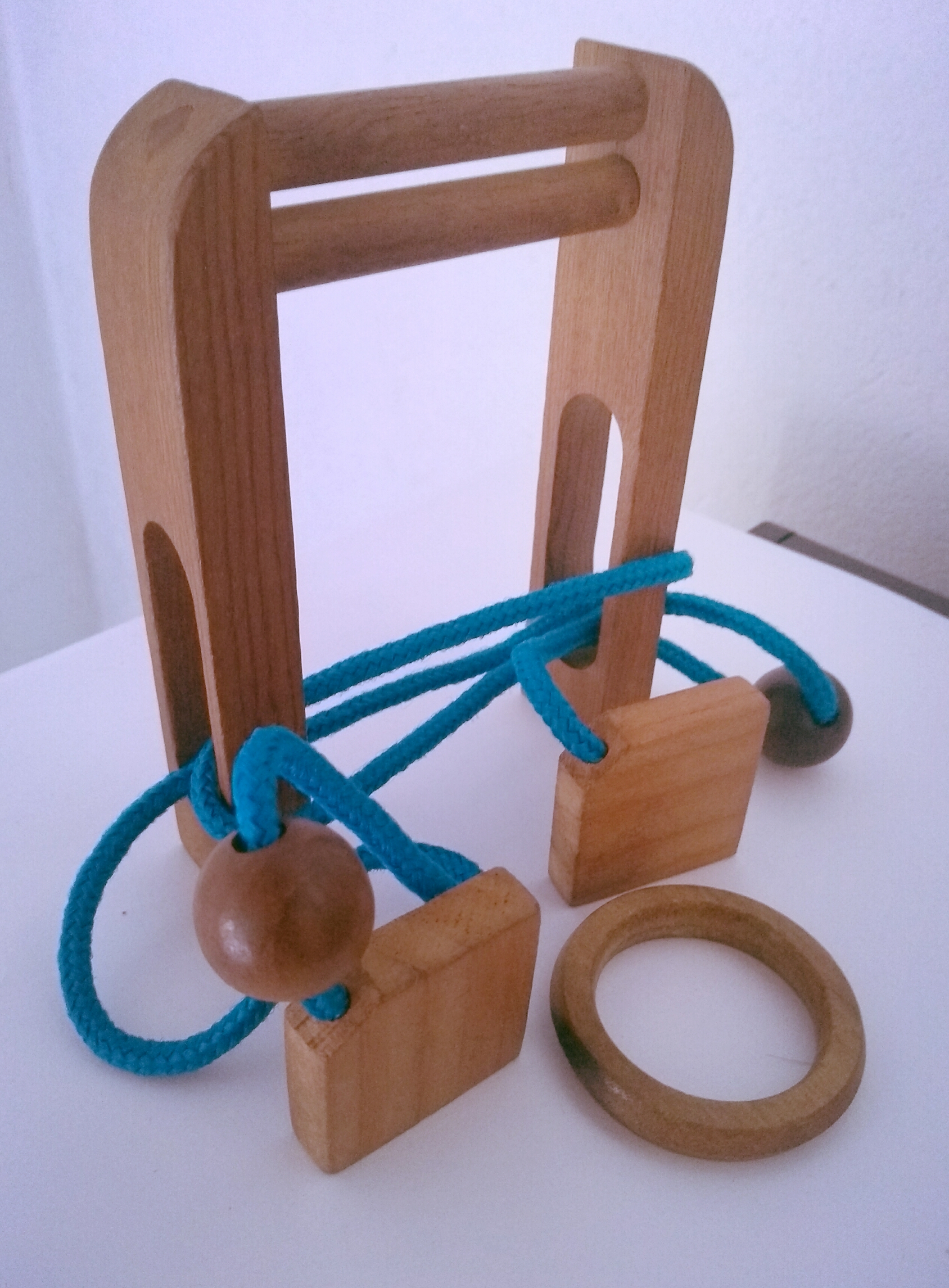}
	\caption{The goal of the Rope Ladder puzzle.}
	\label{fig:RopeLadderPuzzleGoal}
\end{subfigure}
\caption{Initial states and the goal states of the Spatial Puzzles studied in this work.}
\label{fig:SpatialPuzzles}
\end{figure*}

Both puzzles (Figure~\ref{fig:FishermansFollyPuzzle} and Figure~\ref{fig:RopeLadderPuzzle}) are composed of the following elements: a Post with a Hole; a String; a Ring; a pair of Spheres, that are crossed by the String; a pair of Disks, fixed to the two tips of the String. The goal of these puzzles is to free the Ring from the entanglement of objects (Figure~\ref{fig:FishermansFollyPuzzleGoal} and Figure~\ref{fig:RopeLadderPuzzleGoal}) by executing a sequence of actions. In this paper we consider the basic actions defined in~\citep{cabalar2011formalising}, that is, passing an object or a tip of a long object through some hole in a given direction. An example can be the operation: ``{\em pass the Disk through the Post Hole from left to right}.'' 
These passing operations are assumed to be complete: intermediate states in which the object is still partially crossing the hole are disregarded. This assumption was later removed in~\cite{cabalar2016qualitative}, but keeping it is still more convenient for finding the solution, since it avoids the explosion of irrelevant states.

In the Fisherman's Folly, the String crosses the Post in such a way that each Sphere and each Disk stays at one side of the Post. Besides, the Spheres can move through the String, different from the Disks that are fixed to the String's tips. The Spheres are larger than the Post Hole, so they cannot pass through the Post Hole without breaking the puzzle. However, the Spheres can pass through the Ring Hole. The Disks can cross the Post Hole, but they cannot pass through the Ring Hole, whereas the Ring can cross the Post Hole. These interactions between the elements of the puzzle can lead to some complex situations, such as the possibility of winding the String through the Post Hole (or through the Ring Hole) several times, which increases the state space and, consequently, the time needed to find the solution (this will be explored in Section \ref{results}). 

The Rope Ladder puzzle has two Holed Posts and the String crosses these Holes in four different points (see Figure \ref{fig:RopeLadderPuzzle}). In this puzzle there is a similar set of possibilities of interaction as in the Fisherman's Folly, with the addition that in the Rope Ladder puzzle it is possible to pass the Post through both faces of the Ring Hole, in this case, the Post is not fixed to a base (as is the case with the Fisherman's Folly).



With these elements in mind, it is possible to categorize the puzzle's objects following the work by~\cite{cabalar2011formalising} that propose three different categories for the elements: 

\begin{enumerate}
\setlength\itemsep{1em}

\item [1:] \emph{Regular (or simple) Objects}: Disk1, Disk2, Base.

\item [2:] \emph{Objects with Holes}: Ring, Sphere1, Sphere2, PostHole1, PostHole2.

\item [3:] \emph{Long Objects}: String, Post1, Post2.

\end{enumerate}





Since the goal of this work is to build an autonomous agent capable of dealing with these puzzles, it was necessary to find a way for this agent to interact with the domain. To this end the present work uses an Oracle (a simulation of the puzzle) that, given an action executed in a state, it returns the description of the resulting state. This Oracle is the set of predicates describing state transitions included in the PROLOG planner from~\citep{cabalar2011formalising} that uses Iterative Deepening as the search strategy. Although this planner can also be used to solve the Fisherman's Folly puzzle, it is unable to generate a complete solution to more complex (e.g. Rope Ladder), non-deterministic or non-stationary puzzles. As we shall see further in this paper, the heuristic version of oASP(MDP) is capable of handling these domains. It is worth pointing out that the output of oASP(MDP) is a policy, not a plan\footnote{A plan is a sequence of actions that needs to be followed for a goal to be achieved (it is strictly crucial the correct execution of the plan, otherwise a re-plan is needed if something goes wrong, such as the agent going to a different state). A policy is the mapping between state/action, in which there are several state/action pairs that can be executed in order to take the agent to the goal (with no need to find another policy if the agent is in a different state than the optimal states expected).}.

\subsection{Answer Set Programming (ASP)}
Answer Set Programming (ASP)~\citep{eiter2009answer} is a declarative logic programming language that facilitates Non-Monotonic reasoning and it has been successfully used to solve NP-Complete problems, such as the Traveler Salesman Problem~\citep{Hoffman2013}. Besides, it is designed to model and solve problems that deal with commonsense reasoning, such as spatial puzzles~\citep{ThiagoBRACIS2018} or spatial non-monotonic reasoning~\citep{walkega2017non}.

According to ~\cite{eiter2009answer}, ASP presents some advantages that justifies its use, such as: the possibility to define which solutions are more desirable than others using a quality criterion, making it suitable for domains involving preference manipulation; the capability to work with missing information; and the so-called \emph{choice rules}, which in practice, allows for the mapping between one input and several outputs.
One important feature of ASP is that relies on a declarative semantics based on the definition of \emph{stable models}~\citep{lifschitz2008answer}. \cite{gelfond1988stable} define a stable model as ``a possible set of beliefs the agent has, taking into consideration the premises of the program''. 

~\cite{lifschitz2008answer} defines an ASP program as a set of rules of the form: 
\begin{equation} \label{eq:ASPRule}
	\texttt{A :-  L1, L2,..., Ln}  
\end{equation}
where {\tt A} is an atom, the rule \emph{head}, and  {\tt L1,...,Ln} is the rule \emph{body} consisting of literals, that is, either atoms {\tt B} or their default negation {\tt not B}. The symbol  ``{\tt :-}'' can be read as ``if'' and represents a (backwards) logical implication so that \eqref{eq:ASPRule} stands for $\texttt{L1} \wedge \dots \wedge \texttt{Ln} \to \texttt{A}$.
The algorithm applied in this paper uses ASP {\em choice rules}, whose effect is non-deterministic and allows one input to have several outputs~\citep{ferreira2018method}. More formally, given one state ``\(s\)'' and an action ``\(a\)'', it is possible to have the states ``\(s1\)'' and ``\(s2\)'' as possible outputs when the action ``\(a\)'' is executed in state ``\(s\)''. This rule can be represented in ASP with the following Formula:
\begin{equation} \label{eq:stateEquationExample}
	\texttt{1 \{\(s1\), \(s2\)\} 1 :-  \(a\), \(s\).}
\end{equation}
that can be read as: given that the action ``\(a\)'' was executed in the state ``\(s\)'', one and only one state (between ``\(s1\)'' and ``\(s2\)'') can be chosen as a consequence of executing this action. 

The premise of this work is that for each state \(s\) \(\in\) \(S\) (a set of possible states of a domain) there is an ASP program that describes the effects of executing an action \(a\) \(\in\) \(A\) (a set of possible actions that can be executed) using choice rules that inform the possible states that can be reached.

 ASP represents constraints as headless rules. This kind of formulae is mainly used to remove possible solutions that violate some rules. In the context of this work, these constraints are related to the description of the forbidden states (the ones that cannot be physically achieved), actions (that cannot be physically executed) and state/actions pairs (actions that cannot be physically executed in a specific state). 
 ASP is used in this work to find a set of states \(S\) of a Markov Decision Process (MDP), and then, to find all answer sets for each state an agent has permission to visit: all transitions allowed for the state-action $(s,a)$ pair. Also, since ASP can review previously obtained knowledge and describe the transition rules of the domain, it is suitable to represent an MDP in a non-stationary domain. 

\subsection{Reinforcement Learning (RL)} \label{RLSection}
There are some problems that are hard to solve using only pre-defined rules, such as regression, classification and decision making problems in an unknown environment. That is why Machine Learning (ML) techniques are becoming more common lately to solve real world problems. This work uses Reinforcement Learning (RL), which is a method where the learning process happens through the interaction between an agent and the environment.
~\cite{Sutton-Barto2ed} define RL over Markov Decision Process (MDP) that can be expressed as the tuple $\langle S, A, T, R \rangle$, in which:
\begin{enumerate} 
\item [$S$] is the set of possible states in the domain;
\item [$A$] is the set of actions that can be executed by the agent;
\item [$T$] is the transition function that defines the probability of reaching a successor state $s' \in S$ when the agent executes action $a \in A$ starting from state $s \in S$. This function is shown in equation \eqref{eq:transictionFunctionEq};
\item [$R$] is the reward function responsible for providing the reward to the agent, when the agent is in a state $s \in S$ and executes action $a \in A$ to move to successor state $s' \in S$. This function is shown in equation \eqref{eq:rewardFunctoinEq}.
\end{enumerate} 

\vspace*{-4mm}
\begin{equation} \label{eq:transictionFunctionEq}
	\mathrm{Transition} = S \times A \times S \mapsto [0,1]\ 
\end{equation}
\begin{equation} \label{eq:rewardFunctoinEq}
	\mathrm{Reward} = S \times A \times S \mapsto Re
\end{equation}

Besides this, \cite{Sutton-Barto2ed} point out that an MDP assumes a first-order Markov property, which states that state $s$ contains all the information needed by the agent to decide the next action $a$ to be executed.

In RL there are two main entities, the Agent and the Environment. The Agent is responsible for learning and making decisions (executing actions), and the place (or world) in which the Agent executes these actions is part of the Environment.~\cite{Sutton-Barto2ed} define the RL framework of interaction as follows:
\begin{enumerate} 
\item First, the Agent chooses an action \(a\) \(\in\) \(A\) at an instant \(t\) to be executed in the Environment in a state \(s\) \(\in\) \(S\); 
\item Then, the Environment answers the Agent with the next state \(s'\) \(\in\) \(S\) at the next instant \(t_{+1}\) together with the corresponding reward (a numerical value that describes the value of the action executed). 
\item Finally, the Agent updates the value of executing this action based on the received reward, using a formula defined by the specific RL algorithm being used.
\end{enumerate} 

Even though we found a formalism to deal with decision making problems, in which \emph{a priori} information about the Transition and Reward functions are not needed, and defined how the RL interaction works, it is still necessary to specify how to obtain an optimal policy in this case. This paper uses the model-free off-policy algorithm called Q-Learning~\citep{Sutton-Barto2ed}.
The Q-Learning algorithm can learn the optimal policy following the RL framework of interaction defined above. To learn this policy, the algorithm uses an action-value function $Q(s,a)$ that is updated at the end of each interaction between the Agent and the Environment by using equation \eqref{eq:QLearningEq}.
\begin{eqnarray}
	Q_{t+1}(s_t,a_t) & \gets & Q_t(s_t,a_t) \label{eq:QLearningEq}\\
    & & + \alpha \cdot (r + \gamma \cdot \max_{a \in A}  Q_t(s_{t+1},a) - Q_t(s_t,a_t) ) \nonumber
\end{eqnarray}
where \(\alpha\) and \(\gamma\) represent the learning rate and the discount factor, respectively.
These interactions happen several times (defined by the user), and at the end of all of them, the Q-Learning agent is expected to have learned an optimal policy for the problem.

One important point to highlight is that at the beginning of the interaction the Q-Learning agent has no knowledge about which action to choose, the choice is random. This lack of initial knowledge can lead to a poor performance initially for the agent, so it is possible to think in a method that can be used to accelerate this learning procedure, reusing previous obtained knowledge to do this. That is where the work of~\cite{bianchi2008accelerating} becomes relevant for this paper.~\cite{bianchi2008accelerating} present a new approach to the Q-Learning process, the Heuristically Accelerated Q-Learning (HAQL), with the possibility of adding heuristics to guide the learning process when the agent has previous knowledge about the task. 
The heuristics guide the learning process by helping the agent to choose an action in the action selection phase, this is the only difference between the HAQL and the traditional version of Q-Learning, considering that the choice selection is using the \(\epsilon\)-greedy strategy\footnote{A strategy in which the agent has a probability (1 - \(\epsilon\)) of choosing an action that maximizes the value of the policy and a probability (\(\epsilon\)) of choosing a random action.}. This difference can be seen in the following equation for the choice of actions:
\begin{equation} \label{eq:actionChoiceRule}
	\pi(s)=\begin{cases}
        argmax_{a \in A} \{\hat{Q}(s,a) + \xi H(s,a)^\beta\}, & \text{if $q \leq p$}.\\
    random(A), & \text{otherwise},
  \end{cases}
\end{equation}
where $H(s,a)$ is the heuristic function that guides the choice of action, $\xi$ and $\beta$ are parameters that control the influence of the heuristic function, \textit{q} is a random value between 0 and 1, \textit{p} is a value between 0 and 1, responsible for defining the exploitation/exploration trade-off in the Q-Learning algorithm, $random(A)$ chooses some random action $a \in A$. 

With the definition of RL and MDP in mind, we can define how these concepts fit into our work. The environment is represented by the PROLOG program used as Oracle, since the Q-Learning agent can ask the Oracle the consequence of executing an action in a given state. Besides, Q-Learning deals with learning through interaction, which is necessary for the domains in this work, although Q-Learning does not need to know about $T$ and $R$, it still needs to know about the set of actions \(A\) and the set of states \(S\). This justifies the combination with ASP, since it can be used to describe an MDP, to reason about actions and to build the state space in an online manner. In the next section, we present the oASP(MDP) algorithm that combines these two tools.

\subsection{Online Answer Set Programming for Markov Decision Process}
\label{oaspmdp}

\cite{ferreira2018method} define the Online ASP for MDP (oASP(MDP)) as an algorithm that combines the concepts described previously. Taking advantages from ASP with logic programming, MDP with the formalism of a decision making problem, and RL with Q-Learning. ASP is used to describe the MDP and makes it possible to search for answer sets by going through the state space, while the Q-Learning algorithm is used to approximate the action-value function $Q(s,a)$ of this MDP. 

Combining these concepts, we may represent each transition $t(s,a,s')$ as a choice rule:
\begin{equation} \label{eq:ASPEqStateTransition}
	\texttt{1 \{$s'$, $s2'$, $s3'$\} 1 :- $a$, $s$,} 
\end{equation}
every new transition (state discovered, e.g. $s'$, $s2'$ and $s3'$) is added to the choice rule.
This description can be applied to all action and all states in the domain~\citep{ferreira2018method}. So, having the description of possible transitions for each action and each state that has been visited in a given instant as a logic program, an ASP solver can be used to obtain a set of observable states and actions. Finally, through the interactions between the Q-Learning agent and the environment, the agent is able to define the action-value function $Q(s,a)$. 


\begin{algorithm}[ht!]
    \TitleOfAlgo{oASP(MDP)}
    \textbf{Input:} The set of actions \(A\), an action-value function approximation method \(M\) and a number of episodes \(n\).\;
    \textbf{Output:} The approximated \(Q(s,a)\) function.
\DontPrintSemicolon
    \BlankLine
    Initialize the set of observed states \(S = \varnothing\)\;
    \While{number of episodes performed is less than \(n\)}{
        \Repeat{the end of the episode}{
            Observe the current state \(s\)\;
                \eIf{\(s \not \in S\)}{
                    Add \(s\) to the set of states \(S\).\;\label{alg:ifline}
                    Choose and execute a random action \(a \in A\).\;
                    Observe the future state \(s'\).\;
                    Update the logic program for state \(s\) by adding the observed transition in the choice rule.\;
                    Update the description of \(Q(s,a)\) by finding every answer set for each state \(s\) added to \(S\) in this episode.\;
                }{\label{alg:elseline}
                    Choose an action \(a \in A\) as defined by \(M\).\;
                    Execute the chosen action \(a\).\;
                    Observe the future state \(s'\).\;
                }
            Update the value of \(Q(s,a)\) as defined by \(M\).\;\label{alg:update}
            Update the current state \(s \gets s'\).\;
                
    }
}
    \caption{The oASP(MDP) Algorithm created by \cite{ferreira2018method}.}
\label{alg:oaspmdp}
\end{algorithm}

Algorithm \ref{alg:oaspmdp} presents the pseudo-code for oASP(MDP). First, the algorithm receives three different parameters (line 2, algorithm \ref{alg:oaspmdp}) as input: the set of possible actions the agent can initially execute in the environment\footnote{It is worth mentioning that this set of actions can suffer change while the interaction happens, since oASP(MDP) allows for the revision of knowledge.}; then it receives the RL method \(M\) that is used to approximate the action-value function $Q(s,a)$; finally, the algorithm receives the number of episodes \(n\) to be executed. After this initialization, the algorithm defines the set of observed states to empty (line 4, algorithm \ref{alg:oaspmdp}), since this set is built while the interaction happens (online). The biggest advantage of the oASP(MDP) algorithm, compared to traditional Q-Learning, is that even though the construction of the sets \(A\) and \(S\) is done in an online manner, the RL method \(M\) is still capable of approximating the action-value function $Q(s,a)$ and returning the Q-Table, which defines that the optimal policy is followed if the actions with higher values are executed~\citep{ferreira2018method}.

After these first steps, the algorithm starts a loop (that is repeated \(n\) times, lines 5-22). In each repetition of this loop, the oASP(MDP) observes the current state \(s\), and then it can take two different actions (depending if the current state \(s\) is in the set of observed states \(S\) or not) present in \(A\). So, if \(s\) \(\not \in \) \(S\) (line 8, algorithm \ref{alg:oaspmdp}), then \(s\) is added to \(S\) and a random action in set \(A\) is executed. From this first interaction, it is already possible to add the observed transition for this state and action with the choice rule. To finalize this flow of execution, the description of the action-value function Q(s,a) is updated. 

On the other hand, if \(s\) \(\in \) \(S\) (line 14, algorithm \ref{alg:oaspmdp}), then the action is chosen by the RL method \(M\), with no update of the choice rule describing the transition, since this description was already done when the state was first added to set \(S\).

Finally, after these two conditions are met, the update of the values in the action-value function $Q(s,a)$ happens accordingly to the RL method \(M\). The current state \(s\) is also updated, now the current state is the successor state \(s'\) that was observed when the action was executed.


Regarding the changes in the set of actions \(A\), if the action executed by the agent in the environment was one that is physically impossible to be executed (e.g. trying to pass the Sphere1 through the PostHole1), the environment informs this to the agent, that does not take this action into consideration. Another change that can happen is related to the set of possible actions per state. For example, if an action is physically impossible to be executed only for a specific state, then the environment will also convey that to the agent, which can in its turn not take this action into consideration in that specific state. 

Since oASP(MDP) is capable of dealing with changing sets of actions and states, this algorithm becomes very interesting to distinct kinds of problems that have a Non-Stationary\footnote{A Non-Stationary domain is the one in which the environment can change while the interaction is happening.} nature. Different from Q-Learning, that needs to re-initialize the obtained values each time a change happens in the environment, oASP(MDP) can still use and apply the previously learned knowledge even when a change in the environment happens.

The next section presents the application of oASP(MDP) to the spatial puzzle domains. Additionally, we also present the main contribution of this paper: the use of heuristics in oASP(MDP).


\section{Heuristic oASP(MDP) (HoASP(MDP))}
\label{method}
This section presents the adaptation of the oASP(MDP) algorithm to work with heuristics: the HoASP(MDP) algorithm. 
Heuristics are taken into consideration in the instant of the action choice (line 15 of Algorithm \ref{alg:oaspmdp}), which leads to a change in how the original oASP(MDP) algorithm works, so for a better understanding of the applied changes, a demonstration of the HoASP(MDP) algorithm when dealing with the Fisherman's Folly puzzle domain is presented next.

HoASP(MDP) is initialized with the set of Actions \(A\), a learning method \(M\) and the number of episodes \(n\). After that, the agent initializes the set of observed states \(S\) to empty, which is the starting point of the current episode. Next, the agent verifies if the initial state \(s0\) (Figure \ref{fig:FishermansFollyPuzzle}) is in the set \(S\) (line 6, algorithm \ref{alg:oaspmdp}), since in the beginning of the learning procedure the set \(S\) is empty, the agent has to execute a random action (line 10,  algorithm \ref{alg:oaspmdp}), for example, {\em pass the Disk1 through the PostHole1 from right to left}. Following the action execution, the agent goes to the successor state \(s1\), whereas \(s0\) is added to \(S\) and the agent receives the corresponding reward. Now the agent has knowledge about a transition in the domain, that can be translated to a choice rule (line 12, algorithm \ref{alg:oaspmdp}) in ASP as:
\begin{equation*}\label{transitionRuleExample}
	\texttt{1 \{\(s1\)\} 1 :-  \(a\)(ExecutedAction), \(s0\);}
\end{equation*}
where \texttt{ExecutedAction} is the action {\em pass the Disk1 through the PostHole1 from right to left} executed by the agent. Each state \(s\) \(\in \) \(S\) has an ASP file with all these transitions rules, that is updated each time a new transition in that state is detected. Now that ASP has the transition rule, it can find all answer sets for that state and update the description for the action-value function $Q(s,a)$ for that specific state. 

The agent is now in \(s1\), all the steps above are repeated in the same way. However, the following situation can happen: assuming the randomly chosen action was to undo what was done in state \(s0\), executing the action {\em pass the Disk1 through the PostHole1 from left to right}, this leads the agent to \(s0\) again. Since the agent is back in \(s0\), and \(s0\) \(\in \) \(S\), then another HoASP(MDP) flow of execution is followed. The agent now is going to use the learning method \(M\) (line 13, algorithm \ref{alg:oaspmdp}) to choose the action to be executed, and it is also at this instant that the heuristic guiding process takes place. Initially there is the mapping between the state in the heuristic task and the original task (the one that is the focus of study) and then, the agent chooses an action to be executed using equation \eqref{eq:actionChoiceRule}.

In the end of each episode, the received reward is used to update the values for the action-value function $Q(s,a)$, which, after several episodes, is the Q-Table that contains the optimal policy~\citep{Sutton-Barto2ed}.

Regarding the learned ASP rules, each visited state has an ASP file with the learned transitions (for each action executed, there is the consequence of executing an action in a given state), which is updated each time the agent visits a new state. For the Deterministic and Non-Stationary cases, the Formula bellow represents the ASP rules:
\begin{equation*}
    \texttt{1 \{\(s1\)\} 1 :-  \(a\), \(s0\);}
\end{equation*}
in which \(s1\) is the consequence of executing \(a\) in state \(s0\).

Each state can have a maximum of either 20 or 28 rules, depending on the puzzle considered, 20 rules if the puzzle is the Fisherman's Folly and 28 rules if the puzzle is the Rope Ladder. The numbers of rules are related to the number of possible actions that can be executed in the puzzle. 
The number of maximum rules for each state is also either 20 or 28, depending on the puzzle. Besides these transition rules, there is another file, for each state, with the restrictions representing the forbidden actions for a specific state:
\begin{equation*}
\label{formulaRestrictedAction}
    \texttt{:-  \(a\), \(s0\).}
\end{equation*}
where \(a\) is an action from set \(A\) that is impossible to be executed in the physical puzzle when the agent is in \(s0\). Thus, each visited state has two ASP files, one representing the transitions rules and the other with the impossible actions. The last set of rules that are learned by ASP is the forbidden actions for all states, with the Formula:
\begin{equation*}
\label{formulaRestrictedAction}
    \texttt{:-  \(a\).}
\end{equation*}
\(a\) is an action in the set \(A\) that cannot be executed in any state of the puzzle, such as trying to cross the Sphere1 through the PostHole1.


\subsection{Heuristic oASP(MDP) to Solve Spatial Puzzles}
After defining the general principles of the HoASP(MDP) problem solving method, this section describes how this method was adapted to solve the spatial puzzles (of increasing complexity) considered in this paper, with a number of domain versions: Simplified puzzles (relaxed versions of the spatial puzzles), Non-Deterministic (the spatial puzzles with non-determinism when an action is executed) and Non-Stationary (the spatial puzzles with changes in the domain while the interaction happens). 

This work uses two different approaches to obtain domain heuristics, which depend on the puzzle considered. 
In the Deterministic and Non-Deterministic versions of the Fisherman's Folly puzzle, an admissible heuristic was obtained from a simplification of the puzzle solved using the original oASP(MDP). The solution obtained (Q-Table) was used as heuristics to accelerate the learning process for the original puzzle. The Fisherman's Folly simplification kept the same configuration and relations of objects as in the original puzzle. The simplification was a constraint imposed on the number of windings of the String through the Post Hole. In this setting, the String cannot wind through the Post Hole more than twice. 

The solution for the simplification of the Rope Ladder puzzle was also used as heuristics, but there are differences to consider since a distinct strategy was applied to obtain these heuristics. In the Simplified Rope Ladder, the String is initially crossing the two Post Holes only once, which is enough to make the puzzle and the optimal policy simpler. Another point of adaptation, regarding the Simplified Rope Ladder, is the necessity of creating a function that maps a state in the Simplified Rope Ladder to a state in the Original Rope Ladder. Since the set of actions on both puzzles were the same, this mapping function had to match the actions that lead the agent to a certain state in the Simplified puzzle to the actions that lead the agent to the corresponding state in the Original puzzle.


With respect to the actions' formalization in the algorithm, the tuple $\langle \CE, \HE, \HF \rangle$ was used, where:
\begin{enumerate} 
\item $\CE$ is a \emph{Crossing Element}, an element of the puzzle that is going to pass through a hole.
\item $\HE$ is a \emph{Host Element}, an element of the puzzle that hosts a hole.
\item $\HF$ is the \emph{Hole Face}, the face of the hole $\CE$ towards which $\HE$ going to pass. There are two possible faces: positive (+) and negative (-). If the crossing through the hole is from left to right the face is positive, otherwise the face is negative.
\end{enumerate} 

Thus, any given action is the manipulation of a $\CE$ that passes through a $\HE$ toward a direction ($\HF$). This action formalization is general enough to describe the actions for all puzzles (and configurations) considered in this work. Since an action is the combination of a $\CE$, a $\HE$ and a $\HF$, the agent cannot choose an action with the same element as $\CE$ and $\HE$. It is worth mentioning that the agent does not have initial knowledge about which $\CE$ is able to be passed through a $\HE$ due to size or shape constraints: for instance, it does not know whether Sphere1 can pass through the PostHole1, or any other restriction. The agent learns these constraints while it receives rewards that indicates which actions are impossible to be executed. As the number of interactions increases, and the agent learns about the effects of actions, it does not take into consideration impossible actions that are in the set \(A\), which in turn lead the agent to execute fewer impossible actions in the long run. 

In the Fisherman's Folly there is a set of 20 actions over the following elements: 6 $\CE$ elements \{Sphere1, Sphere2, Post, Disk1, Disk2, Ring\}; 2 $\HE$ elements \{PostHole1, Ring\}; and 2 $\HF$ \{Positive, Negative\}. 

The formalization of the actions for the Rope Ladder is very similar to the Fisherman's Folly: the only difference is in the crossing and host elements, that in this case correspond to $\CE$=\{Sphere1, Sphere2, Post1, Post2, Disk1, Disk2, Ring\}, $\HE$= \{PostHole1, PostHole2, Ring\}, leading to 28 possible actions to be chosen by the agent. 

Similar to the approach described by~\cite{santos2016framing}, the states of the puzzles are represented with a list of crossings per each long object in the puzzle.
So, in the Fisherman's Folly, we have a list for the objects being crossed by the String, and another for the objects being crossed by the Post. For example, for the initial state shown in Figure~\ref{fig:FishermansFollyPuzzle}, the lists of crossings can be represented as follows:
\begin{enumerate} 
\setlength \itemsep{1em} 
\item chain(String) = [+Sphere1, +Post, +Sphere2]
\item chain(Post) = [+Ring]
\end{enumerate}
where each list element is some object name from $\HE$ preceded by a hole face sign (the ``exit'' of the crossing). For instance, the crossing +Ring in the Post list means that the Post is crossing the Ring towards the positive face of the Ring Hole.



\section{Experiments}
\label{experiments}
The goal of this section is to evaluate how distinct RL approaches work with variants of two spatial puzzles, Fisherman's Folly and Rope Ladder, in different configuration settings. 
Four RL algorithms were applied to the puzzles, as described in Sections \ref{background} and \ref{method}. Q-Learning \citep{Sutton-Barto2ed}, the original oASP(MDP) \citep{ferreira2018method}, Heuristically Accelerated Q-Learning (HAQL) \citep{bianchi2008accelerating} and Heuristic oASP(MDP) (HoASP(MDP)), proposed in this paper.
For the comparison between these methods, some parameters were defined for all four algorithms in all experiments: the discount factor equals to 0.9; \(\epsilon\) (trade-off between exploitation and exploration) was fixed to 0.1 until the 4000th episode, after that the value was decreased with a rate of 0.01 at each 250 episodes, down to the value of 0.03; the learning rate was fixed to 0.2; the heuristic control value was 0.25. The rewards were defined as follows: -100 to impossible actions (those actions that lead to the same state or that are impossible to be performed on the physical puzzle), 1000 to actions that lead to the goal state and -1 to each action performed by the agent that leads to a different state. The agent can execute 500 actions per episode and the total number of episodes, by trial, is 6000. To produce statistically relevant results, 30 trials were executed in each experiment.

This paper considers the following distinct configurations of the puzzles:

\begin{itemize}
    \item{\textbf{\textit{Simplified Fisherman's Folly:}}}
        In this configuration, the elements of the puzzle are the same as in the original Fisherman's Folly (Figure~\ref{fig:FishermansFollyPuzzle}), but the agent cannot wind the String through the Post Hole (this is the only element with this restriction), which simplifies the state space. When this restriction is applied, it is possible to see how it can affect the puzzle and consequently the time and number of visited states required to solve it. Two algorithms were applied in this domain, the original oASP(MDP) and Q-Learning. 
        
    \item{\textbf{\textit{Original Fisherman's Folly:}}}
        This is the original Fisherman's Folly puzzle, presented in Figure \ref{fig:FishermansFollyPuzzle}. All four RL algorithms assumed in this article were applied to this domain (cf. Sections \ref{background} and \ref{method}). Regarding the algorithms that use heuristics, the Q-Table of the previous puzzle (Simplified Fisherman's Folly) was used to guide the learning procedure. We extend the work of~\cite{ThiagoBRACIS2018} through the application of oASP(MDP) in the Fisherman's Folly with different configurations and exploring a different set of RL parameters.
        
    
    \item{\textbf{\textit{Non-Deterministic Fisherman's Folly:}}}
        In this version, when an agent chooses an action to execute, there is an 80\% chance of that action achieving the expected results, a 10\% chance of the action to be executed, but mistakenly aiming at the opposite hole face and a 10\% chance that the agent executes no action (staying at the same state). The optimal sequence of actions to solve it is the same as the original puzzle. All four RL algorithms described in this article were applied to this domain. As in the original Fisherman's Folly, the solution to the Simplified Fisherman's Folly was used as heuristics in the non-deterministic case.
        
        
    \item{\textbf{\textit{Non-Stationary Disk Fisherman's Folly:}}}
        Although, the work by~\cite{ferreira2017answer} presents the application of the offline ASP(MDP) to Non-Stationary domains, in the present paper, we wanted to evaluate how the oASP(MDP) algorithm deals with environment changes when a previous learned policy can be a  drawback for a new task.
        Thus, in this domain, only the original oASP(MDP) and traditional Q-Learning algorithms were executed, with no application of heuristics. In the beginning, the domain allows for both Disk1 and Disk2 to cross through the Ring, but not through the Post Hole. The change happens after the 2000th episode, when the puzzle becomes the original Fisherman's Folly (where the Disks cannot pass through the Ring).

    
    \item{\textbf{\textit{Original Rope Ladder:}}}
        In terms of complexity, this is one of the most difficult puzzles to solve in the set, since there are more elements connected in a more challenging configuration when compared to the Fisherman's Folly puzzle. The optimal sequence of steps to solve the Rope Ladder has 12 actions. All four RL algorithms described in this article were applied in this domain. Regarding the algorithms that use heuristics, the Q-Table of a simplified version of the Rope Ladder was used to guide the learning procedure. In the simplified version, there is only one crossing between the String and  PostHole1 and PostHole2, which simplifies the puzzle in such a way that becomes feasible to use it as the puzzle to extract the heuristics. Another constraint applied to both configuration of the Rope Ladder puzzle is that the agent cannot pass Disk1 through PostHole2 nor Disk2 through PostHole1 and it cannot wind the String through the Posts's Hole two (or more) consecutive times.
        
\end{itemize}
To evaluate and compare the algorithms considered in this work in the spatial puzzles described above, four metrics were used: the number of steps to solve the puzzle; the accumulated return values; the number of states visited by the agent; and the number of state/actions pairs in the Q-Table obtained during the learning process. The results and discussions of these experiments are presented in the next section. 

\section{Results and Discussion}
\label{results}
This section presents the results of applying the algorithms considered in this work on distinct configurations of spatial puzzles. This facilitates a comparison of Q-Learning and oASP(MDP) with respect to their heuristically accelerated versions. 

\subsection{Simplified Fisherman's Folly (SFF)}
For the Simplified Fisherman's Folly (SFF) puzzle, only two algorithms were executed: oASP(MDP) and Q-Learning, since it is the solution of this domain that provides heuristics to most of the more complex domains described below.\par
Figure~\ref{fig:numberOfStepsSimplifiedFishermans} shows the graph for the Number of Steps to solve the puzzle, it is possible to see that the learning curves of both algorithms are similar, this is due to the fact that the learning algorithm used by oASP(MDP) is Q-Learning, so this similarity is expected.\par

Figure~\ref{fig:returnValuesSimplifiedFishermans} represents the graph for the Accumulated Return Values, although both curves are similar, the return values for the oASP(MDP) algorithm are slightly better at the beginning of the learning process, because oASP(MDP) has the ability to remove impossible actions (the ones that cannot be executed in the puzzle and receive a negative reward with high absolute value), in contrast with Q-Learning that keeps these impossible actions on the Q-Table and can execute them in all visited states. \par

\begin{figure*}[h]
\centering
\begin{subfigure}[t]{1.0\textwidth}
	\centering
	\includegraphics[width=\columnwidth]{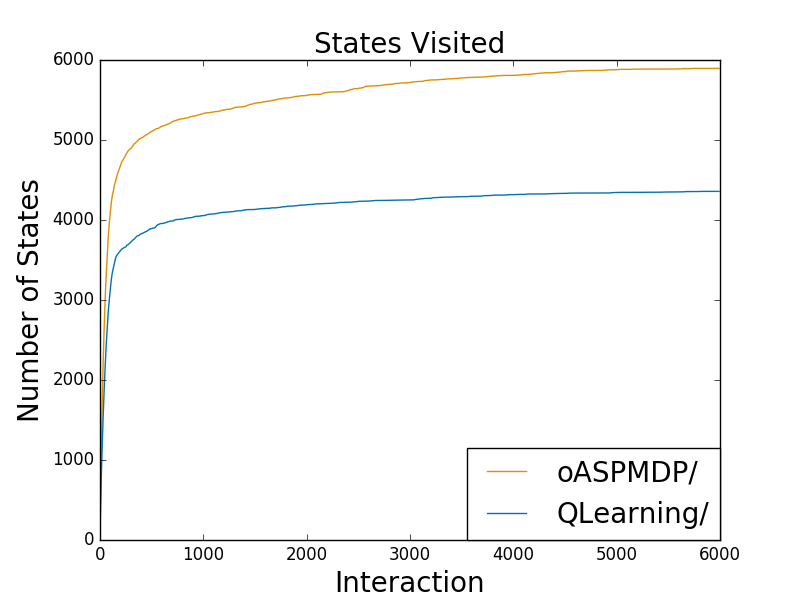}
	\caption{Number of States visited by the Agent.}
	\label{fig:numberOfStatesSimplifiedFishermans}
\end{subfigure}
\\
\begin{subfigure}[t]{1.0\textwidth}
	\centering
	\includegraphics[width=\columnwidth]{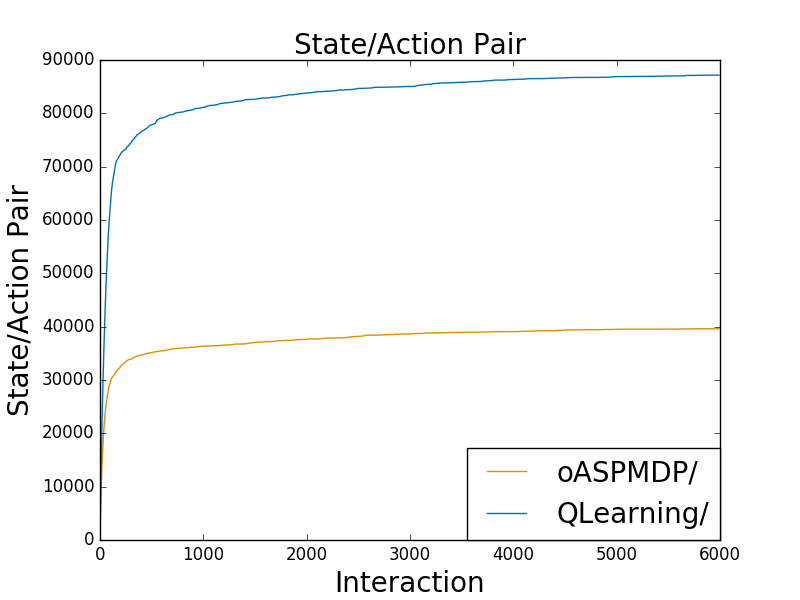}
	\caption{Number of State/Actions pairs.}
	\label{fig:numberOfStateActionsSimplifiedFishermans}
\end{subfigure}
\caption{States results for the Simplified Fisherman's Folly puzzle.}
\label{fig:graphsStateSimplifiedFishermans}
\end{figure*}

Figure~\ref{fig:numberOfStatesSimplifiedFishermans} shows the number of Visited States. The oASP(MDP) agent visits more states because it can remove impossible actions, which leads the agent to execute more valid actions with new states as outcome. On the other hand, Q-Learning keeps these impossible actions, which implies in a low exploration of the environment. \par

Finally, Figure~\ref{fig:numberOfStateActionsSimplifiedFishermans} shows the number of State/Actions pairs present in the Q-Table, showing that Q-Learning accumulated more pairs. This graph shows a distinction with respect to the graph of Visited States (Figure \ref{fig:numberOfStatesSimplifiedFishermans}) because all the actions are already in the Q-Table once the state is visited. Since the Q-Learning algorithm does not revise its knowledge about actions, it does not remove impossible actions from the Q-Table, it only assigns negative rewards to them.

\subsection{Original Fisherman's Folly (OFF)}
For the Original Fisherman's Folly (OFF), all 4 algorithms were executed. Figure~\ref{fig:numberOfStepsCompleteFishermans} shows the graph with the Number of Steps to solve the puzzle. We can also see that using the solution of a simplified version as heuristics to a more complex domain accelerates the learning process from the beginning, this is due to the fact that the solution of the SFF puzzle is part of the solution for the OFF puzzle. The Student's T Test in Figure~\ref{fig:tTestCompleteFishermans} shows that these two curves present differences that are statistically relevant. \par

Figure~\ref{fig:returnValuesCompleteFishermans} represents the graph for the Accumulated Return Values, which also shows a better performance for the heuristic algorithms compared to their non-heuristic versions.\par

\begin{figure*}[h]
\centering
\begin{subfigure}[t]{1.0\textwidth}
	\centering
	\includegraphics[width=\columnwidth]{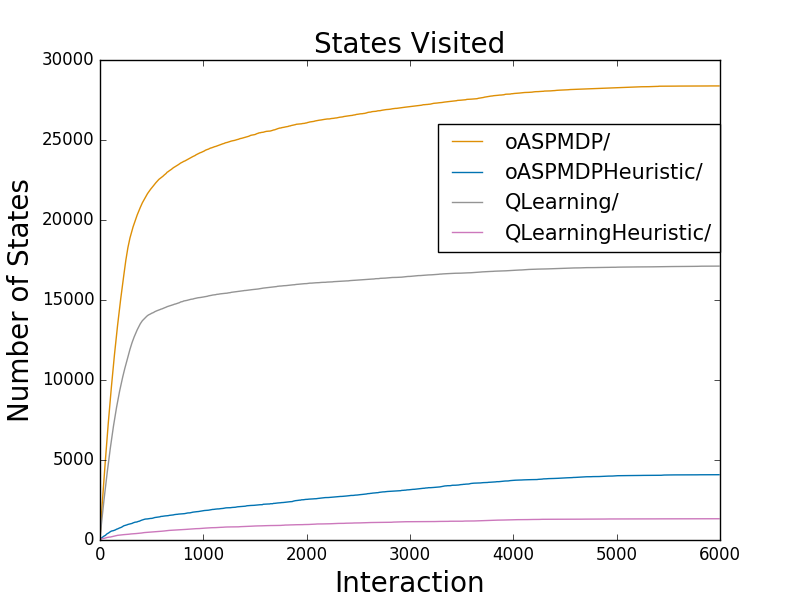}
	\caption{Number of States visited by the Agent.}
	\label{fig:numberOfStatesCompleteFishermans}
\end{subfigure}
\\
\begin{subfigure}[t]{1.0\textwidth}
	\centering
	\includegraphics[width=\columnwidth]{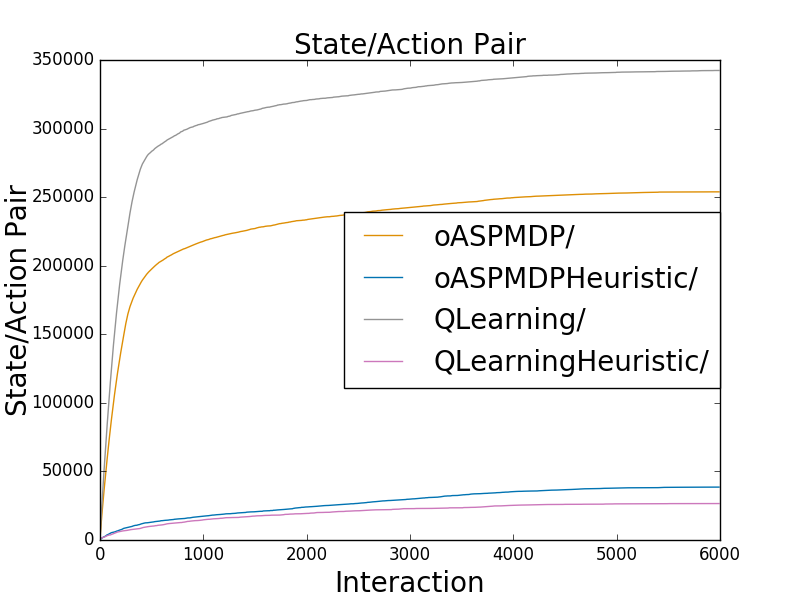}
	\caption{Number of State/Actions pairs.}
	\label{fig:numberOfStateActionsCompleteFishermans}
\end{subfigure}
\caption{States results for the Original Fisherman's Folly puzzle.}
\label{fig:graphsStateCompleteFishermans}
\end{figure*}

Figure~\ref{fig:numberOfStatesCompleteFishermans} shows the number of Visited States by the agent, oASP(MDP) visits more states than the other 3 algorithms. Comparing the number of interactions that is necessary to learn the optimal policy (Figure~\ref{fig:numberOfStepsCompleteFishermans}) and the number of Visited States, it is possible to see that the possibility of winding the String through the Post Hole increases the complexity to solve the OFF puzzle, in contrast with the SFF puzzle (Figure~\ref{fig:numberOfStepsSimplifiedFishermans} and Figure~\ref{fig:numberOfStatesSimplifiedFishermans}). Considering the State Space of the SFF puzzle, the oASP(MDP) agent visits around 6,000 states, whereas in the OFF puzzle the oASP(MDP) agent visits around 25,000 states. Heuristic algorithms explore a smaller portion of the State Space than their non-heuristic versions, this is because the heuristics are responsible for guiding the learning process from the beginning, indicating to the agent which actions to execute and, consequently, which states to visit, accelerating these algorithms to find solutions.\par

Finally, Figure~\ref{fig:numberOfStateActionsCompleteFishermans} shows the number of State/Actions pairs present in the Q-Table. The Q-Learning is the algorithm with the highest number of State/Actions pairs.


\subsection{Non-Deterministic Fisherman's Folly (NDFF)}
The Number of Steps to solve the Non-Deterministic Fisherman's Folly (NDFF) is shown in Figure~\ref{fig:numberOfStepsCompleteNonDeterministicFishermans}. It is possible to see that, at the beginning of the learning process, the heuristic algorithms perform better than their related non-heuristic versions. Student's T test in Figure~\ref{fig:tTestCompleteNonDeterministicFishermans} indicates that this difference is statistically relevant. \par

Figure~\ref{fig:returnValuesCompleteNonDeterministicFishermans} shows the graph for the Accumulated Return Values, in which the heuristic algorithms present higher return values at the beginning of the process. Another point to consider is that the oASP(MDP) has higher return values than Q-Learning in the beginning of the interactions, but as the number of interactions increases, this difference declines until they achieve the same level. \par

\begin{figure*}[h]
\centering
\begin{subfigure}[t]{1.0\textwidth}
	\centering
	\includegraphics[width=\columnwidth]{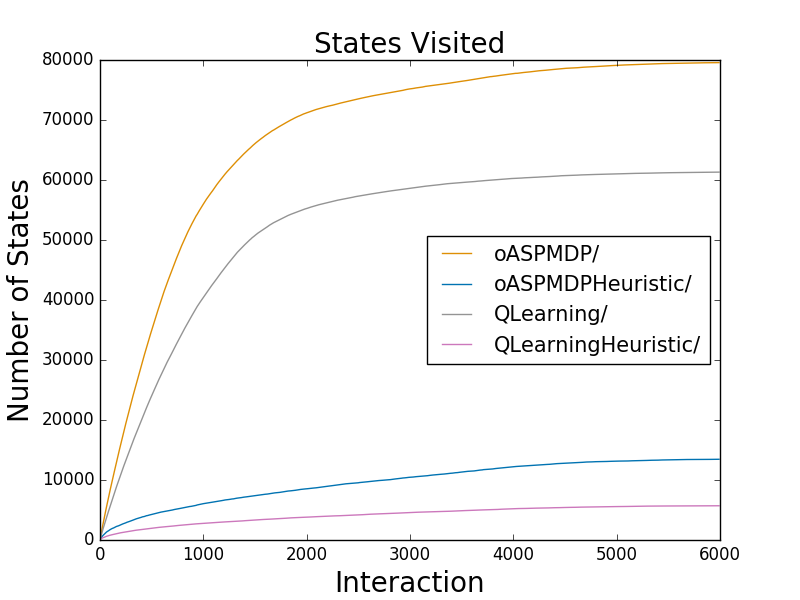}
	\caption{Number of States visited by the Agent.}
	\label{fig:numberOfStatesCompleteNonDeterministicFishermans}
\end{subfigure}
\\
\begin{subfigure}[t]{1.0\textwidth}
	\centering
	\includegraphics[width=\columnwidth]{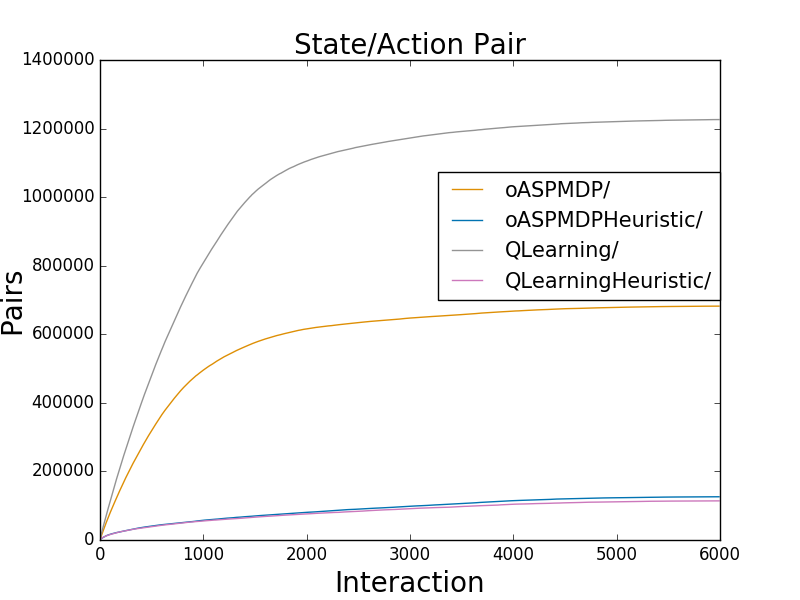}
	\caption{Number of State/Actions pairs.}
	\label{fig:numberOfStateActionsCompleteNonDeterministicFishermans}
\end{subfigure}
\caption{States results for the Non-Deterministic Fisherman's Folly puzzle.}
\label{fig:graphsStateCompleteNonDeterministicFishermans}
\end{figure*}

The graph in Figure~\ref{fig:numberOfStatesCompleteNonDeterministicFishermans} shows that oASP(MDP) explores more valid states than all the other algorithms, while Figure~\ref{fig:numberOfStateActionsCompleteNonDeterministicFishermans} shows that Q-Learning has more State/Actions pairs in the Q-Table. Since, in this configuration, it is also possible to wind the String through the Post Hole, the oASP(MDP) agent explores a higher number of states than in any other configuration of this puzzle (Figure~\ref{fig:numberOfStatesCompleteNonDeterministicFishermans}), visiting around 80,000 states. Thus, the algorithms take longer to learn (Figure~\ref{fig:numberOfStepsCompleteNonDeterministicFishermans}). \par

Although this is a non-deterministic problem, the algorithms were still able to use heuristics to accelerate the learning process, because the action choice procedure depends on the heuristics influence only at the moment the agent is actually choosing the action to be executed. Thus, even though these actions have non-deterministic effects, in the overall learning process, this non-determinism does not present a complication in terms of heuristics choice.


\subsection{Non-Stationary Disk Fisherman's Folly (NSFF)}
Considering the results for the Non-Stationary Disk Fisherman's Folly (NSFF) puzzle, Figure~\ref{fig:numberOfStepsNonStationaryFishermansDisk} shows that both algorithms have a similar performance until the change in the environment happens (at interaction 2000). After that, Q-Learning performs better than oASP(MDP), since the latter does not achieve the optimal policy. This is a statistically relevant difference as demonstrated by the Student's T Test in Figure~\ref{fig:tTestNonStationaryFishermansDisk}. The way the change happens in the environment impacts the performance of the oASP(MDP) agent, since the oASP(MDP) agent learns a policy that can be a drawback in the new version of the environment (after interaction 2,000). In other words: first, the oASP(MDP) agent learns an optimal policy to solve the problem before the change occurs; second, after the change is detected, its Q-Table is not re-initialized, so the optimal policy, previously learned, influences the behaviour in the new environment, which is not the optimal policy for the problem after the change. The oASP(MDP) agent does not receive enough negative reward in order to overturn these actions, which leads the agent to execute some actions that are not optimal (after the change), consequently converging to a sub-optimal policy for the second part of the learning process.\par 

Although there is a difference in the Number of Steps (Figure~\ref{fig:numberOfStepsNonStationaryFishermansDisk}) to solve the puzzle after the change occurs, this difference cannot be observed in the Accumulated Return Values graph (Figure~\ref{fig:returnValuesNonStationaryFishermansDisk}) because the reward at each step is small (-1) when compared to the reward for achieving the goal (+1,000).

\begin{figure*}[h]
\centering
\begin{subfigure}[t]{1.0\textwidth}
	\centering
	\includegraphics[width=\columnwidth]{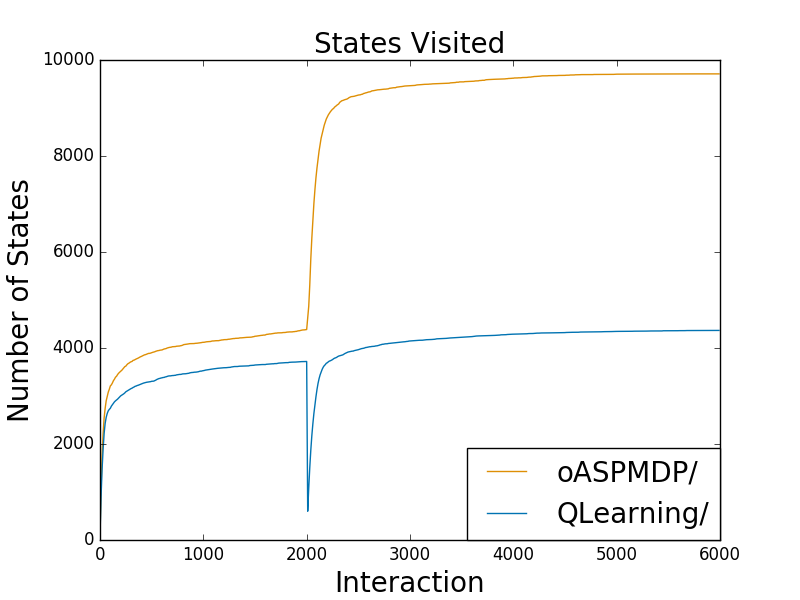}
	\caption{Number of States visited by the Agent.}
	\label{fig:numberOfStatesNonStationaryFishermansDisk}
\end{subfigure}
\\
\begin{subfigure}[t]{1.0\textwidth}
	\centering
	\includegraphics[width=\columnwidth]{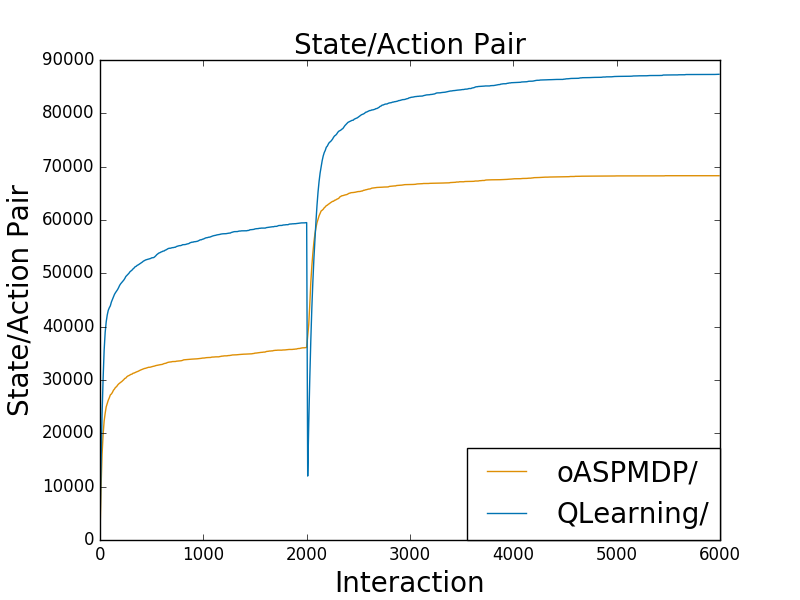}
	\caption{Number of State/Actions pairs.}
	\label{fig:numberOfStateActionsNonStationaryFishermansDisk}
\end{subfigure}
\caption{States results for the Non-Stationary Disk Fisherman's Folly puzzle.}
\label{fig:graphsStateNonStationaryFishermansDisk}
\end{figure*}

Regarding the behaviour related to the Visited States (Figure~\ref{fig:numberOfStatesNonStationaryFishermansDisk}), the oASP(MDP) keeps the Q-Table, even though there is a change in the environment, while Q-Learning needs to re-initialize its values. Figure~\ref{fig:numberOfStateActionsNonStationaryFishermansDisk} shows that Q-Learning has more State/Actions pairs, even though it re-initializes the Q-Table. The number of Visited States and State/Actions pairs just grow for the oASP(MDP) agent because the Q-Table is not re-initialized, in contrast with its Q-Learning agent.

\subsection{Rope Ladder}  
\begin{figure*}[h]
\centering
\begin{subfigure}[t]{0.9\textwidth}
	\centering
	\includegraphics[width=\columnwidth]{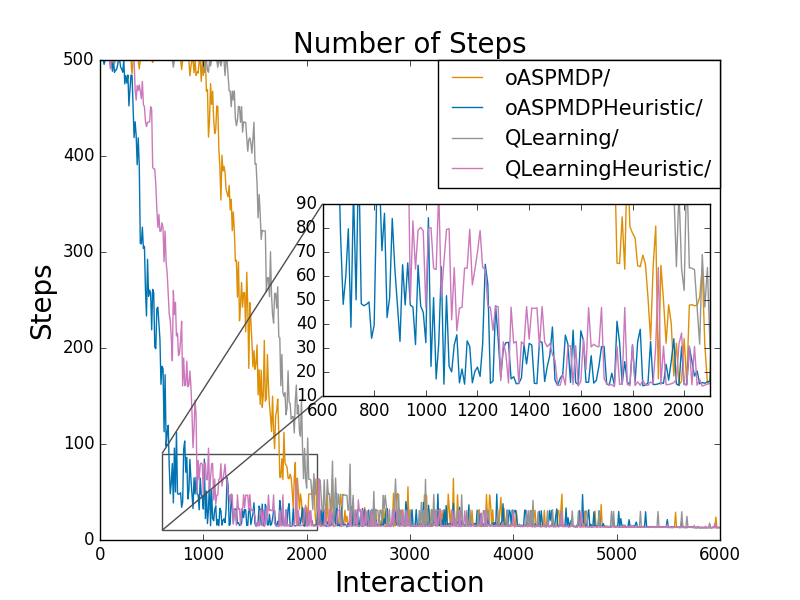}
	\caption{Number of Steps to solve the puzzle.}
	\label{fig:numberOfStepsRopeLadder}
\end{subfigure}
\\
\begin{subfigure}[t]{0.9\textwidth}
	\centering
	\includegraphics[width=\columnwidth]{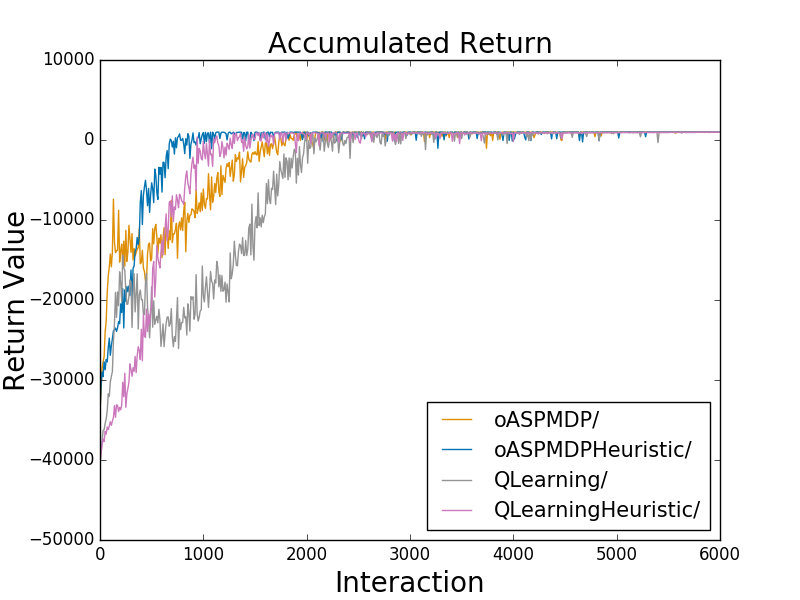}
	\caption{Total accumulated Return received per episode.}
	\label{fig:returnValuesRopeLadder}
\end{subfigure}
\\
\begin{subfigure}[t]{0.8\textwidth}
	\centering
	\includegraphics[width=\columnwidth]{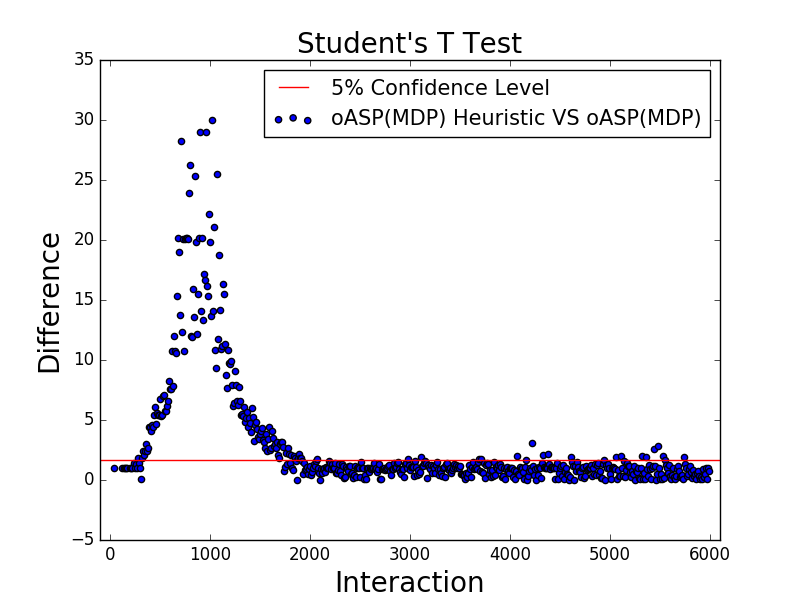}
	\caption{T Test comparing the oASP(MDP) with Heuristic and the traditional oASP(MDP).}
	\label{fig:tTestRopeLadder}
\end{subfigure}
\caption{Number of Steps and Return results for the Rope Ladder puzzle.}
\label{fig:graphsRopeLadder}
\end{figure*}

\begin{figure*}[h]
\centering
\begin{subfigure}[t]{0.8\textwidth}
	\centering
	\includegraphics[width=\columnwidth]{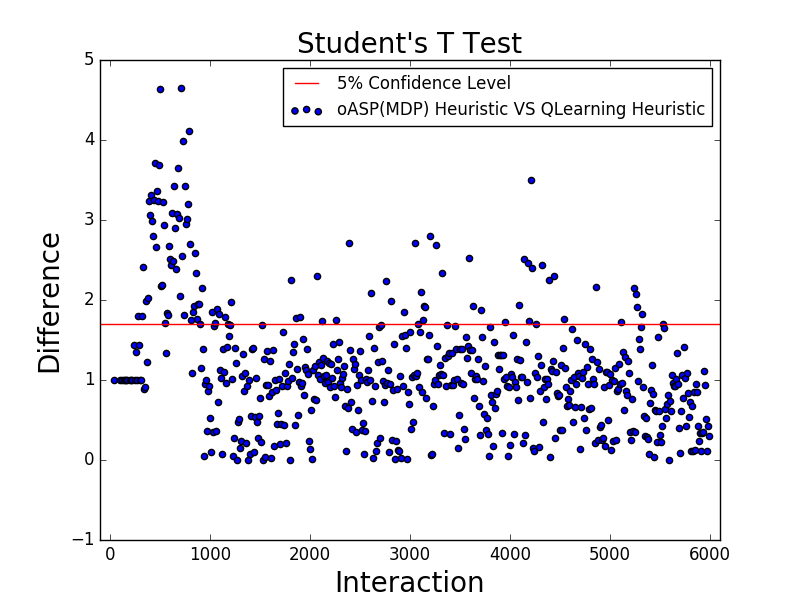}
	\caption{T Test comparing the oASP(MDP) with Heuristic and HAQL.}
	\label{fig:tTestHeuriscAcceleratedRopeLadder}
\end{subfigure}
\\
\begin{subfigure}[t]{0.9\textwidth}
	\centering
	\includegraphics[width=\columnwidth]{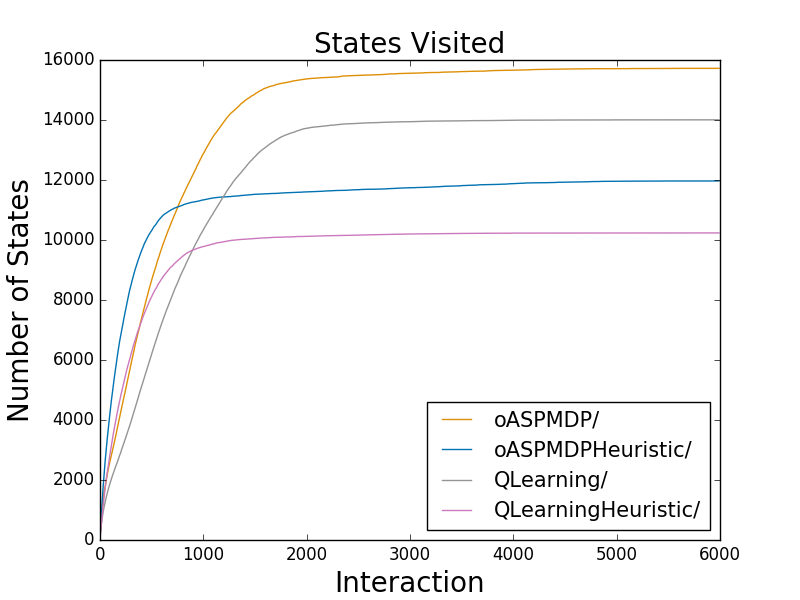}
	\caption{Number of States visited by the Agent.}
	\label{fig:numberOfStatesRopeLadder}
\end{subfigure}
\\
\begin{subfigure}[t]{0.9\textwidth}
	\centering
	\includegraphics[width=\columnwidth]{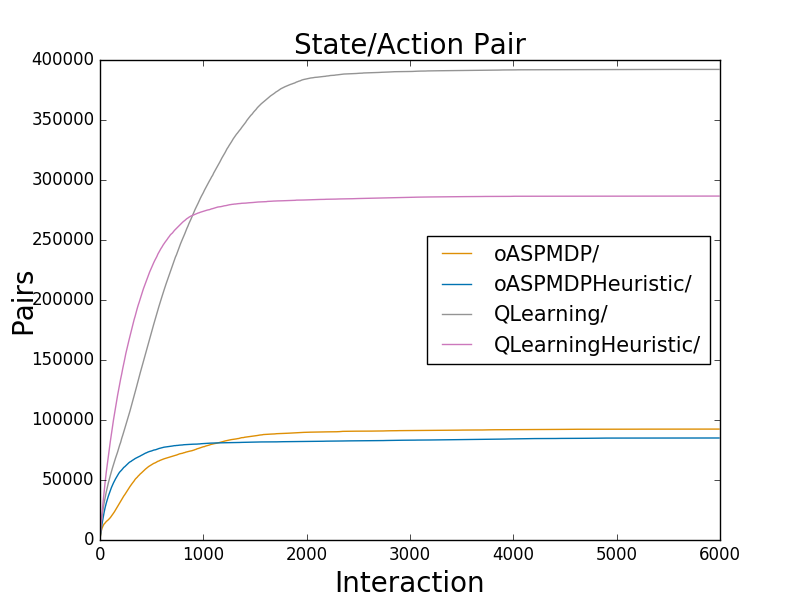}
	\caption{Number of State/Actions pairs.}
	\label{fig:numberOfStateActionsRopeLadder}
\end{subfigure}
\caption{States results for the Rope Ladder puzzle.}
\label{fig:graphsStateRopeLadder}
\end{figure*}

Results related to the Number of Steps to solve the Rope Ladder puzzle are shown in Figure~\ref{fig:numberOfStepsRopeLadder}, that presents how the learning curve for the heuristic algorithms are different (learning faster the optimal policy, showing that heuristics greatly accelerate the learning process) with respect to those of their non-heuristic versions. This difference is statistically relevant as shown by the Student's T Test (Figure~\ref{fig:tTestRopeLadder}). Besides, the HoASP(MDP) also presents for a brief number of interactions (around 300 interactions) a better performance than the Heuristically Accelerated Q-Learning, the Student's T Test supporting this difference is in Figure~\ref{fig:tTestHeuriscAcceleratedRopeLadder}. \par

Figure~\ref{fig:returnValuesRopeLadder} presents the graph for the Accumulated Return Values, it is possible to note how these values change along time: at the beginning of the interaction, the non-heuristic algorithms presents higher return values than the heuristic algorithms, but this relation changes as the number of interactions increases, since the heuristic algorithms start to get higher return values with time. \par

One important observation that can be made about this domain is in relation to the Accumulated Return Values graph (Figure~\ref{fig:returnValuesRopeLadder}) and the number of Visited States (Figure~\ref{fig:numberOfStatesRopeLadder}), it is possible to see that at the beginning of the learning process (in the zoomed part of the graph), the algorithms that do not use heuristics have a higher return value than the algorithms that use heuristics. This happens because the number of Visited States does not increase in the same proportion as the reward value for the non-heuristic algorithms. Thus, the non-heuristic algorithms visit the same states more often, they take longer to explore the state space of the domain, which leads to the execution of fewer impossible actions, since they have already received rewards indicating which actions are better in those known states. This is true only at the beginning of the interactions, after that, the heuristic algorithms explore the state space receiving higher rewards. The return values increase for the heuristic algorithms, while the non-heuristic algorithms take longer to get to the same results. 

Finally, the graph for the State/Actions pairs in Figure~\ref{fig:numberOfStateActionsRopeLadder} shows that Q-Learning has a higher number of pairs than the other algorithms.

\begin{figure*}[h]
\centering
\begin{subfigure}[t]{0.9\textwidth}
	\centering
	\includegraphics[width=\columnwidth]{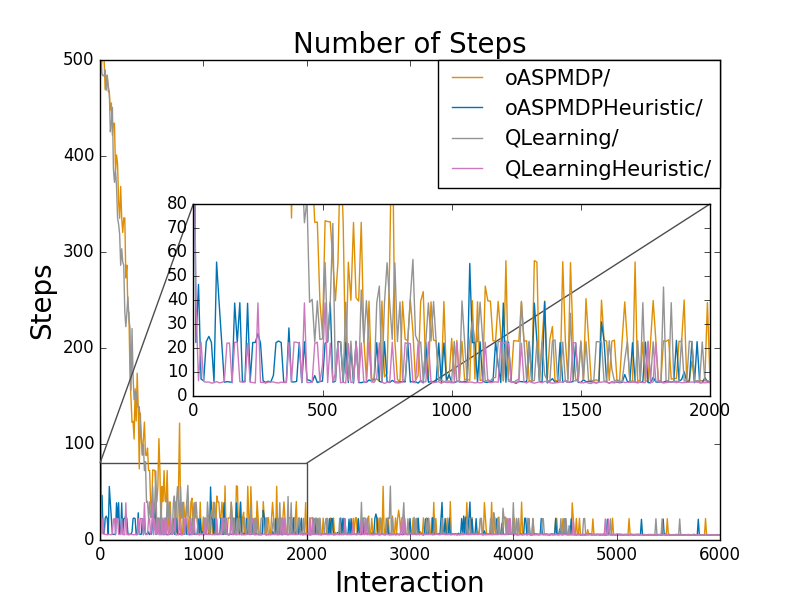}
	\caption{Number of Steps to solve the puzzle.}
	\label{fig:numberOfStepsCompleteFishermans}
\end{subfigure}
\\
\begin{subfigure}[t]{0.9\textwidth}
	\centering
	\includegraphics[width=\columnwidth]{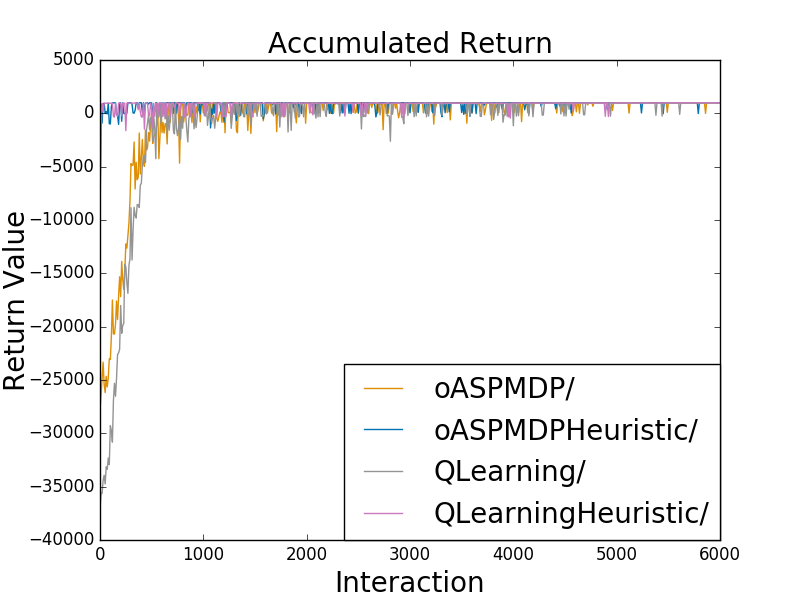}
	\caption{Total accumulated Return received per episode.}
	\label{fig:returnValuesCompleteFishermans}
\end{subfigure}
\\
\begin{subfigure}[t]{0.8\textwidth}
	\centering
	\includegraphics[width=\columnwidth]{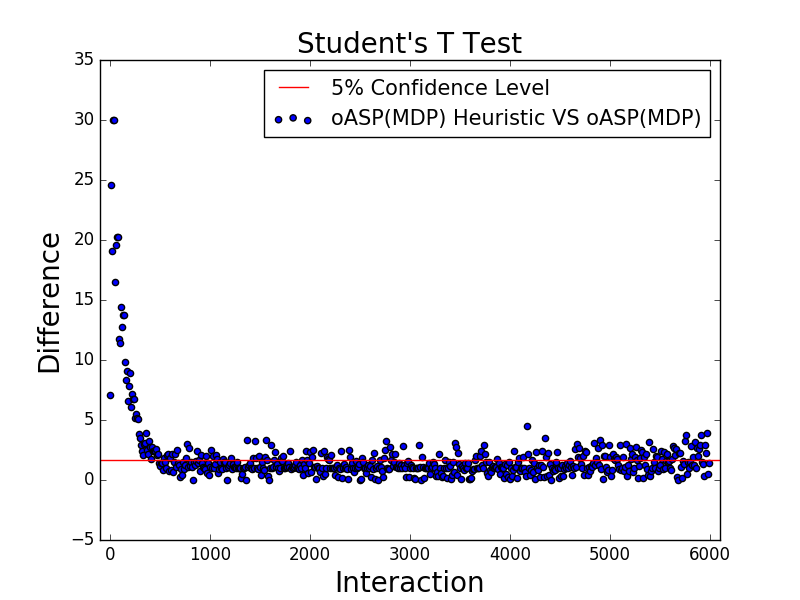}
	\caption{T Test comparing the oASP(MDP) with Heuristic and the traditional oASP(MDP) .}
	\label{fig:tTestCompleteFishermans}
\end{subfigure}
\caption{Number of Steps and Return results for the Original Fisherman's Folly puzzle.}
\label{fig:graphsCompleteFishermans}
\end{figure*}

\begin{figure*}[h]
\centering
\begin{subfigure}[t]{0.9\textwidth}
	\centering
	\includegraphics[width=\columnwidth]{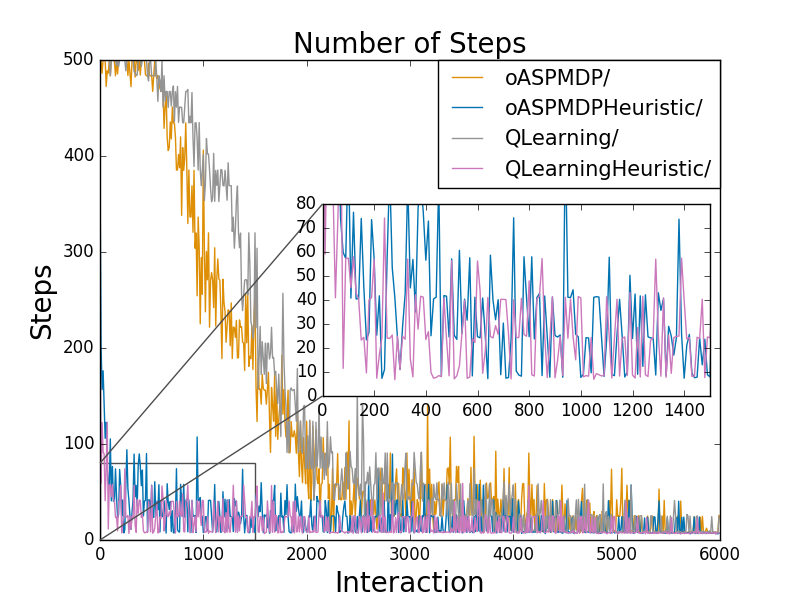}
	\caption{Number of Steps to solve the puzzle.}
	\label{fig:numberOfStepsCompleteNonDeterministicFishermans}
\end{subfigure}
\\
\begin{subfigure}[t]{0.9\textwidth}
	\centering
	\includegraphics[width=\columnwidth]{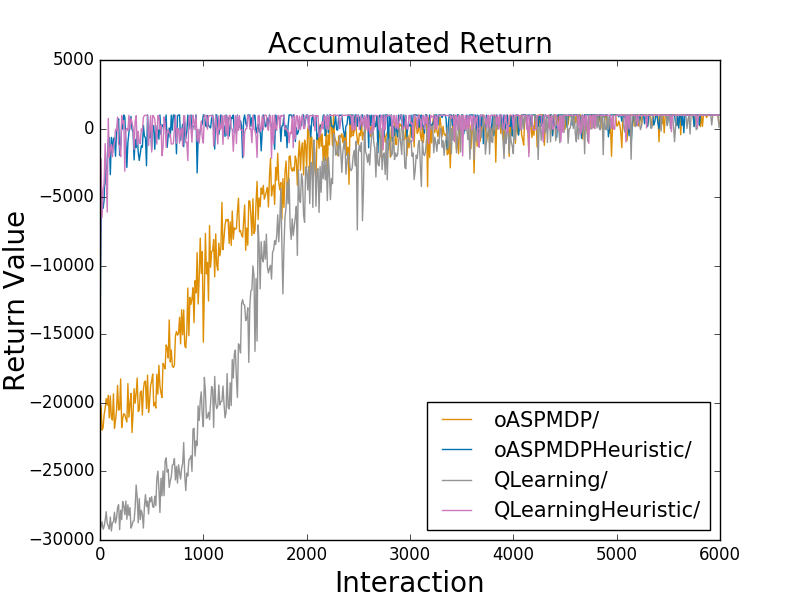}
	\caption{Total accumulated Return received per episode.}
	\label{fig:returnValuesCompleteNonDeterministicFishermans}
\end{subfigure}
\\
\begin{subfigure}[t]{0.8\textwidth}
	\centering
	\includegraphics[width=\columnwidth]{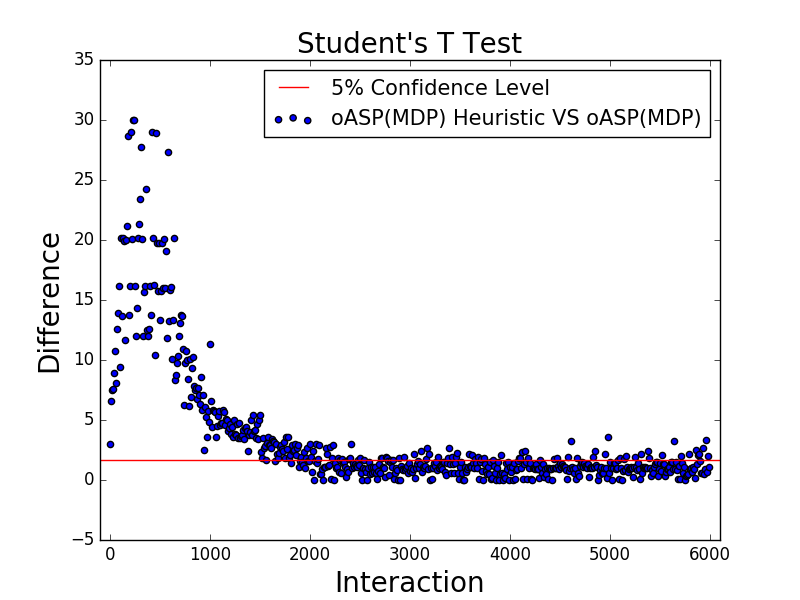}
	\caption{T Test comparing the oASP(MDP) with Heuristic and the traditional oASP(MDP).}
	\label{fig:tTestCompleteNonDeterministicFishermans}
\end{subfigure}
\caption{Number of Steps and Return results for the Non-Deterministic Fisherman's Folly puzzle.}
\label{fig:graphsCompleteNonDeterministicFishermans}
\end{figure*}

\begin{figure*}[h]
\centering
\begin{subfigure}[t]{0.9\textwidth}
	\centering
	\includegraphics[width=\columnwidth]{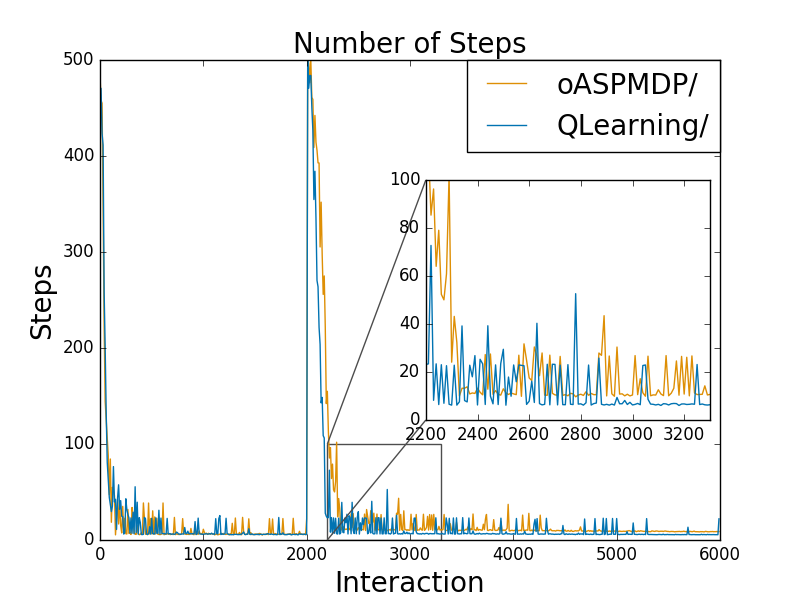}
	\caption{Number of Steps to solve the puzzle.}
	\label{fig:numberOfStepsNonStationaryFishermansDisk}
\end{subfigure}
\\
\begin{subfigure}[t]{0.9\textwidth}
	\centering
	\includegraphics[width=\columnwidth]{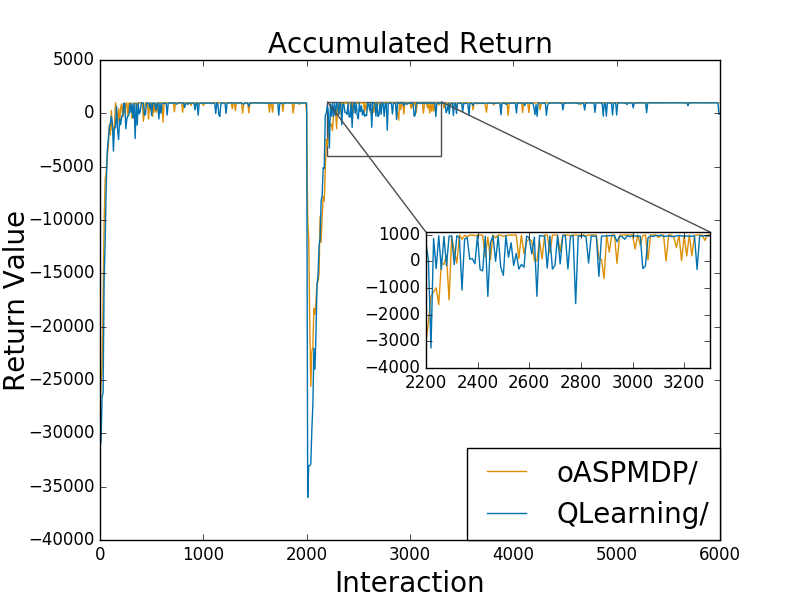}
	\caption{Total accumulated Return received per episode.}
	\label{fig:returnValuesNonStationaryFishermansDisk}
\end{subfigure}
\\
\begin{subfigure}[t]{0.8\textwidth}
	\centering
	\includegraphics[width=\columnwidth]{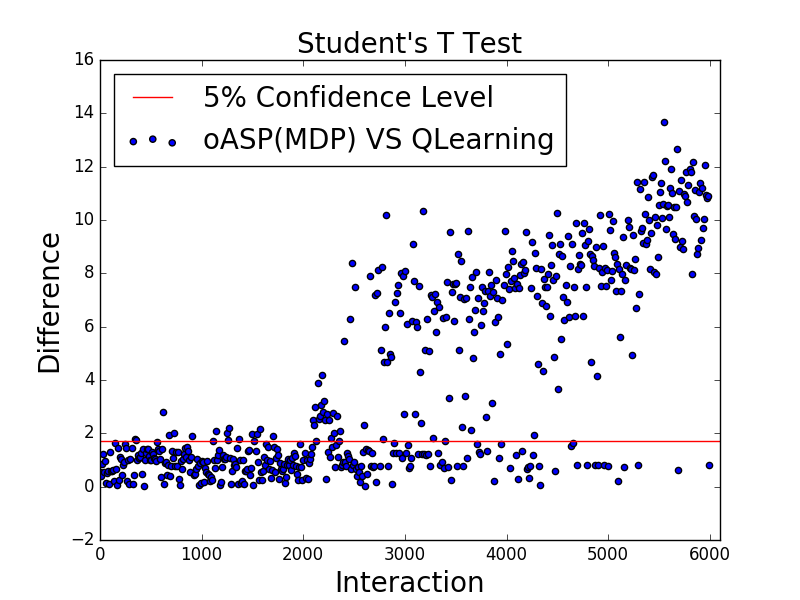}
	\caption{T Test comparing the oASP(MDP) with Heuristic and the traditional oASP(MDP) .}
	\label{fig:tTestNonStationaryFishermansDisk}
\end{subfigure}
\caption{Number of Steps and Return results for the Non-Stationary Disk Fisherman's Folly puzzle.}
\label{fig:graphsNonStationaryFishermansDisk}
\end{figure*}

There are also some general observations that can be made about the Number of Steps graph for all puzzles configurations (Figures \ref{fig:numberOfStepsSimplifiedFishermans}, \ref{fig:numberOfStepsCompleteFishermans}, \ref{fig:numberOfStepsCompleteNonDeterministicFishermans}, \ref{fig:numberOfStepsNonStationaryFishermansDisk} and \ref{fig:numberOfStepsRopeLadder}). It seems that when the learning process is stable (after it finds the optimal solution), and since the agent still has an exploration rate, the learning curves appear to be stable in a certain number of steps (above the optimal policy). This behaviour happens because the agent, in one of the trials, does not find the optimal solution due to the exploration, which leads the agent to execute all 500 actions in that episode not finding the goal state. Thus, the average value calculated for that episode is above the value for the optimal solution.\par
    
Another point is that heuristics were not used in the non-stationary cases because these domains, after the change occurs, were too distinct from their initial configurations (making the heuristic useless). We also wanted to evaluate how oASP(MDP) differs from Q-Learning in these cases because real world applications can have a non-stationary nature and we had already shown (in the results for the OFF, NDFF and Rope Ladder puzzles) how the use of heuristics causes a positive impact in the automated solution for this set of puzzles. \\

The experiments were run on a 3.2GHz Intel Core I7-8700 with 16GB of RAM, in Ubuntu 16.04 LTS. The algorithms were implemented using Python 3.5, using ZeroMQ to provide message exchange between the environment and the agent. {\tt SWI-Prolog} was used as the environment oracle and {\tt clingo} was the ASP engine. The source code for the experiments is available at: \url{https://bitbucket.org/thiagomestrado/journalarticle/src} 

\begin{figure*}[h]
\centering
\begin{subfigure}[t]{1.0\textwidth}
	\centering
	\includegraphics[width=\columnwidth]{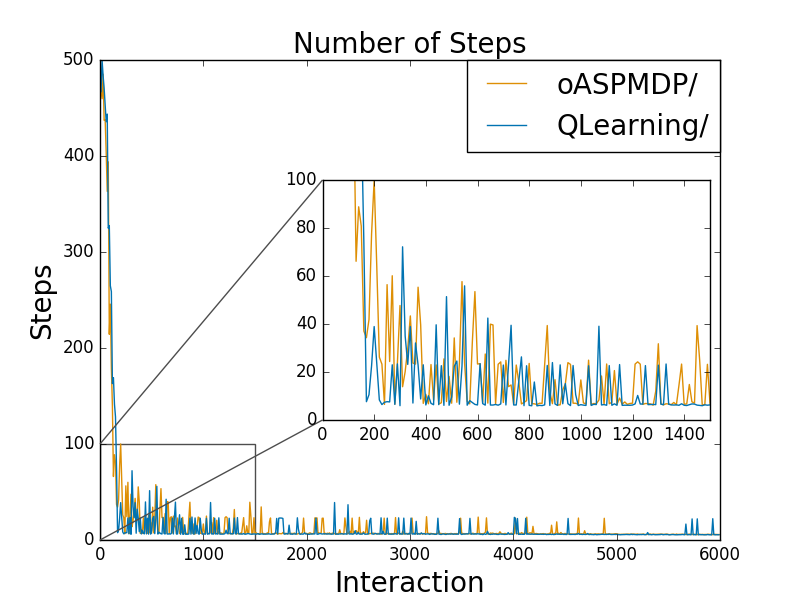}
	\caption{Number of Steps to solve the puzzle.}
	\label{fig:numberOfStepsSimplifiedFishermans}
\end{subfigure}
\begin{subfigure}[t]{1.0\textwidth}
	\centering
	\includegraphics[width=\columnwidth]{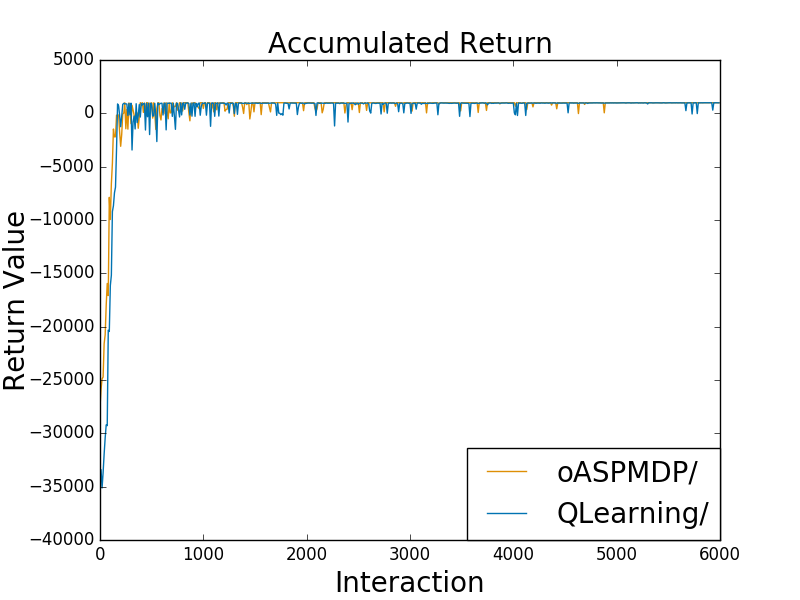}
	\caption{Total accumulated Return received per episode.}
	\label{fig:returnValuesSimplifiedFishermans}
\end{subfigure}
\caption{Number of Steps and Accumulated Return results for the Simplified Fisherman's Folly puzzle.}
\label{fig:graphsSimplifiedFishermans}
\end{figure*}

\section{Literature Review}
This section presents related work to the research described in the present paper. Specifically, we cite literature related to the the combination of Logic and MDP and the use of heuristics in Reinforcement Learning.

In order to find a solution to the Fisherman's Folly and Rope Ladder puzzles, the oASP(MDP) algorithm was used in this paper, an algorithm that combines logic and MDP. Another research that explores a similar combination is described by~\cite{nickles2012system}, that proposes a system (named QASP) that integrates RL (Q-Learning to approximate the action-value function $Q(s,a)$) and ASP, in which the latter is used to present, compute and constrain the set of states and actions. QASP can be applied to non-deterministic problems, but not to non-stationary problems, because it applies ASP after a determined action was chosen and the $Q(s,a)$ function was updated. In contrast, oASP(MDP) generates the \textit{answer sets} before the action-selection phase and before the approximation process of the $Q(s,a)$ function, which makes our algorithm tolerant to changes.\par

~\cite{zhang2015mixed} also present a related approach, combining Answer Set Programming with POMDP, as a framework for a robot in a partially observable environment (real and simulated) executing non-deterministic actions. That approach uses POMDP to describe actions and also to deal with uncertainty about the robot's sensor readings. This is different from the oASP(MDP) investigated in the current paper due to the base formalism used in the domain description. Besides, the work by~\cite{zhang2015mixed} relies on \emph{a priori} planning phase to provide an estimate for the POMDP, while the latter is used to describe the domain and its uncertainty.\par 

The work described by~\cite{LEONETTI2016103} explores the combination of ASP and RL in the DARLING framework. This framework uses ASP in order to represent models and to allow planning and reasoning, while RL is used to make the agent adaptive to the environment. DARLING was tested with a service robot in an office-like environment, allowing the authors to show that, when DARLING is used, the robot can learn tasks faster, improving its performance over time.\par 

~\cite{garnelo2016towards} present an architecture to combine Deep Learning with Symbolic Reasoning, separating it in two main blocks, one formed by a Neural Network (NN), which gives as output a symbolic representation to the other block of the architecture, responsible for the action choice process. In this scenario, Deep Learning is used to find a description of a set of states that can be described as rules for a probabilistic logic program.\par 

The work reported by~\cite{ZMWW17} combines Deep Q-Network with Symbolic Representation, where the latter is used as input and output of the Deep Learning algorithm. The representation in~\cite{ZMWW17} describes spatial relations between two objects and, since there can be a high number of combination of predicates, these relations make the Q-table an interactive model. The predicates representing the states are passed from the environment to the RL agent. The difference with respect to the oASP(MDP) is in the state generation phase, since in oASP(MDP) this phase is the responsibility of the environment, with no use of an intermediate system that maps a set of predicates to a state.\par

~\cite{yang2018peorl} presents a framework that integrates symbolic planning with Hierarchical Reinforcement Learning (HRL), that deals with decision making problems in a non-deterministic domain. The symbolic plans are used to guide the learning procedure, after this step, the learned experience is used again to improve further the provided plan for the HRL algorithm. This approach is define as Planning, Execution, Observation and Reinforcement Learning (PEORL). In this context, the BC language is used to represent common sense knowledge. Results showed that PEORL presents higher return values than traditional versions of RL. \par

The oASP(MDP) algorithm is adapted in this paper to consider heuristics following the ideas developed in previous work. \cite{bianchi2008accelerating} propose the Heuristically Accelerated Reinforcement Learning (HARL) that applies heuristics to accelerate and guide the reinforcement learning procedure. ~\cite{martins2013heuristically} presents a comparative study of three different RL algorithms with their heuristic accelerated versions. The authors show that the use of good heuristics is very useful to guide the action selection phase in RL, when an agent has little knowledge (especially at the beginning of the learning procedure). Moreover, as the interactions happen, the agent presents a better global performance due to the initial acceleration by heuristics. Also, after some number of interactions (depending on the domain) the importance of the heuristics decreases, which leads the algorithms to behave in similar ways. Still exploring this combination,~\cite{morozsHARL2015} applied the HARL approach to the domain of dynamic secondary spectrum-sharing in cellular systems\footnote{A problem that deals with how to share the available spectrum, in the radio networks context, allowing for better voice calls and data transmissions, besides providing a good quality of service to the users.}. Experiments showed that the use of heuristics can help to achieve high control of the sharing patterns in a totally autonomous way. Besides, results also showed that the HARL approaches outperform traditional RL procedures.\par

\cite{bianchi2017heuristically} presents a class of algorithms that use Case Based Reasoning (CBR) as heuristics in a transfer learning setting in order to accelerate the learning process. This approach was applied to two different robotic domains where, first, the RL algorithm is applied to a source task (the task used to extract the heuristics); and then, after the learning stabilizes, the case base is built. The learned cases are transferred to be used in the target task. Similarly, the use of heuristics in the work described in the present paper could be linked to a transfer learning process, since the heuristics are built from simpler versions of the domains. We shall explore this idea further in future work.\par

\label{literatureReview}

\section{Conclusion}
\label{conclusion}
This paper explored the use of heuristics in a method that combines Answer Set Programming with Reinforcement Learning in a Markov Decision Process (oASP(MDP)) modeling a set of spatial puzzles. 
This work considered two base puzzles composed of entanglements involving flexible strings, rigid objects and holes: the Fisherman's Folly and the Rope Ladder, where the latter has more objects submitted to more challenging relations than the former. Experiments were executed on distinct versions of these puzzles, defining deterministic, non-deterministic, and non-stationary domains. These experiments were performed in order to evaluate how distinct algorithms could deal with these domain characteristics. Four algorithms were compared in this study: Q-Learning, Heuristically Accelerated Q-Learning (HAQL), oASP(MDP) and the main contribution of this work, the Heuristic version of oASP(MDP) (HoASP(MDP)). 

The heuristics were obtained from the solution of a relaxed version of the domains and the results show that these heuristics provide important information to guide and accelerate the learning process, with the heuristic accelerated algorithms always outperforming their relative non-heuristic versions. In particular, the oASP(MDP) and HoASP(MDP) algorithms demonstrated to be capable of exploring a larger portion of the valid state space by executing fewer impossible actions than their counterparts (Q-Learning and HAQL, respectively). 

In order to explore further the use of ASP within a Markov decision process, future work will focus on applying ASP to extract general domain rules (constraints, for example) and then apply these rules to reason about the domain, possibly guiding the learning process. In this way, we can have a more general framework for reusing previously learned knowledge. Having a more general description of the state, allows for the representation of rules in ASP that are elaboration tolerant, from a Knowledge Representation point of view.


%

\bibliographystyle{spbasic}      

\end{document}